\PassOptionsToPackage{table}{xcolor}
\documentclass[11pt]{article}
\usepackage{booktabs,multirow,graphicx,float}
\usepackage{agenticlearning-pdflatex}
\usepackage{tabularx}
\usepackage{xspace}
\usepackage{pifont}
\usepackage{tikz}
\usepackage{tcolorbox}
\usetikzlibrary{arrows.meta,positioning}


\definecolor{AwardOrange}{HTML}{C2410C}
\definecolor{MaxAccent}{HTML}{A7444E}
\definecolor{ForestGreen}{HTML}{14532D}
\definecolor{LabPurple}{HTML}{41007F}
\hypersetup{
  colorlinks=true,
  linkcolor=LabBlue,
  citecolor=LabPurple,
  urlcolor=AwardOrange,
  filecolor=ForestGreen,
  anchorcolor=LabDark,
  breaklinks=true
}
\colorlet{controlblue}{LabBlue}
\colorlet{changeorange}{LabRed}
\colorlet{maxorange}{AwardOrange}
\colorlet{successgreen}{ForestGreen}
\colorlet{regressionred}{LabRed}
\colorlet{lightgray}{LabLightGray}
\newcommand{\maxbench}{\textsc{LIBERO-MAX}\xspace}
\newcommand{\cmark}{\textcolor{successgreen}{\ding{51}}}
\newcommand{\xmark}{\textcolor{black!48}{\ding{55}}}
\newcolumntype{Y}{>{\raggedright\arraybackslash}X}
\newcolumntype{Z}{>{\centering\arraybackslash}X}
\newtcolorbox{rqanswerbox}[1]{
  title={#1},
  colback=MaxAccent!5,
  colframe=MaxAccent,
  colbacktitle=MaxAccent,
  coltitle=white,
  fonttitle=\bfseries,
  boxrule=0.7pt,
  arc=2mm,
  outer arc=2mm,
  boxsep=0pt,
  left=7pt,
  right=7pt,
  top=7pt,
  bottom=7pt,
  toptitle=5pt,
  bottomtitle=5pt,
  before skip=8pt,
  after skip=8pt
}
\newcommand{\rqanswer}[2]{\begin{rqanswerbox}{#1}#2\end{rqanswerbox}}

\title{\maxbench:\\ Do Robot Policies Adapt When the World Changes?}

\author{
Yunbei Zhang\textsuperscript{1,*,\textdagger},
Zijian Jin\textsuperscript{2,*},
Yuanzhe Liu\textsuperscript{3},
Janet Wang\textsuperscript{1},
Xilun Zhang\textsuperscript{4},
Yuyou Zhang\textsuperscript{5},\\
Zhenyu Zhang\textsuperscript{4},
Daoan Zhang\textsuperscript{6},
Shuaicheng Niu\textsuperscript{7},
Gen Li\textsuperscript{7},
Jianfei Yang\textsuperscript{7},\\
Jihun Hamm\textsuperscript{1},
Ismini Lourentzou\textsuperscript{3},
Weirui Ye\textsuperscript{8},
Bo Liu\textsuperscript{9},
Peter Stone\textsuperscript{9},
Marco Pavone\textsuperscript{4}
\\[3pt]
\footnotesize
\textsuperscript{1}Tulane University \enspace
\textsuperscript{2}New York University \enspace
\textsuperscript{3}UIUC \enspace
\footnotesize
\textsuperscript{4}Stanford University \enspace
\textsuperscript{5}CMU \enspace
\footnotesize
\textsuperscript{6}University of Rochester \enspace\\
\textsuperscript{7}Nanyang Technological University \enspace
\textsuperscript{8}MIT CSAIL \enspace
\footnotesize \textsuperscript{9}The University of Texas at Austin
}
\date{}

\begin{document}
\shorttitle{LIBERO-MAX}
\shortauthor{Zhang et al.}
\maketitle
\begingroup
\renewcommand{\thefootnote}{\fnsymbol{footnote}}
\footnotetext[1]{Equal contribution.}
\footnotetext[2]{Corresponding author (\textcolor{black}{yzhang111@tulane.edu}).}
\endgroup

\vspace{-7mm}
\begin{center}
\small\textbf{Project website:} \url{https://liberomax.github.io/}
\end{center}
\vspace{-3mm}

\begin{figure}[!h]
    \centering
    \includegraphics[width=\linewidth]{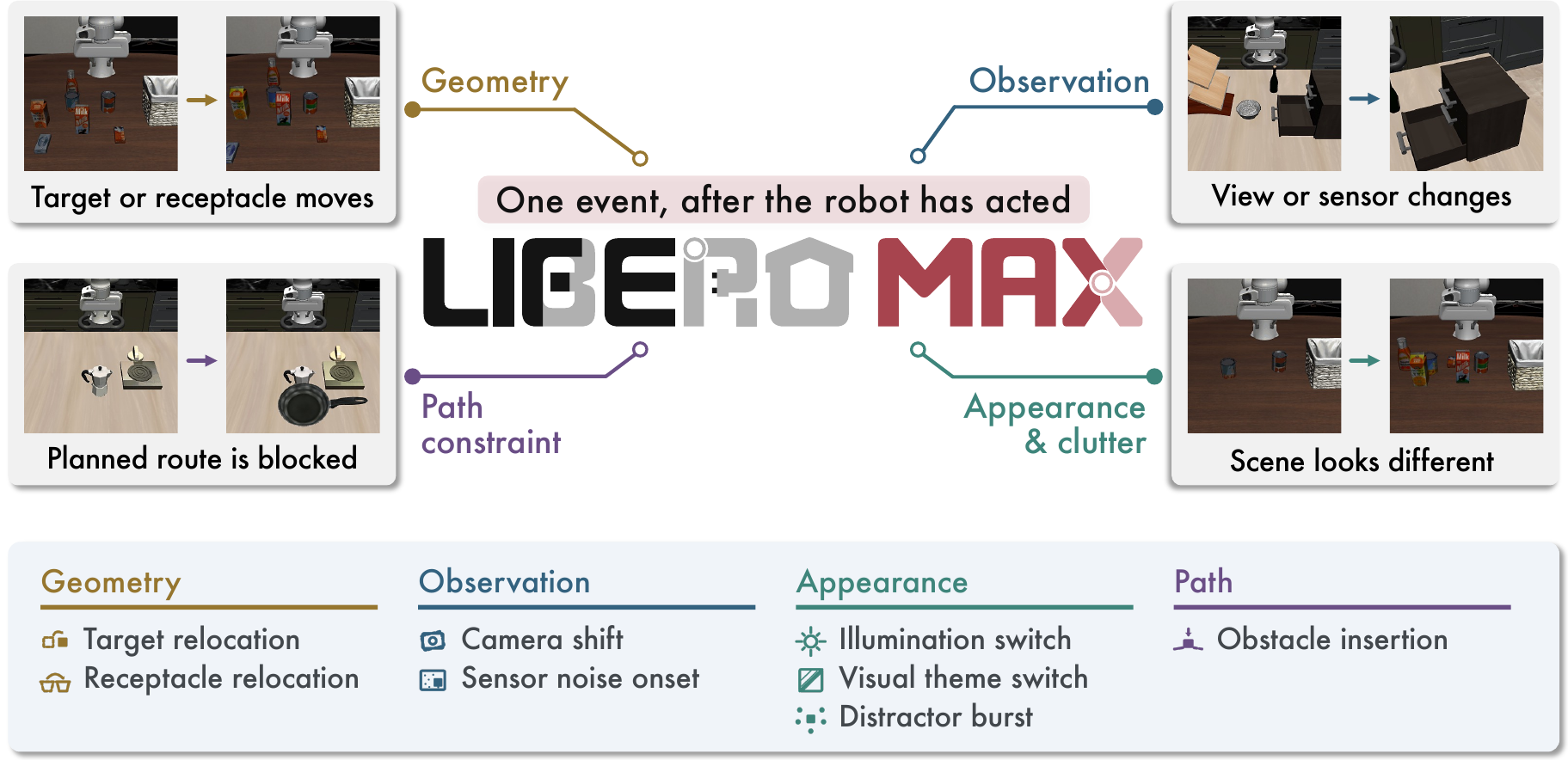}
    \caption{
   \textbf{\maxbench evaluates whether robot policies can complete tasks when the world changes during execution.} Each case pairs two rollouts with identical pre-event actions: Base continues without an added event, while Dynamic receives one mid-task change. Before-and-after images illustrate four event families: geometry, observation, appearance and clutter, and path constraints. The lower panel lists all eight event types.
    }
    \label{fig:overview}
    \vspace{-2mm}
\end{figure}

\begin{abstract}
Robots must often continue a task after a target moves, the viewpoint shifts, or an obstacle appears, even though their earlier observations and committed actions reflect the previous scene. Many simulation robustness benchmarks fix external conditions at reset, leaving this temporal challenge underexamined.
We introduce \maxbench, a benchmark of 8,000 paired cases spanning eight types of changes to geometry, observations, appearance, clutter, and paths. Each pair compares task execution with and without a mid-task event, holding the task, initial state, policy seed, and pre-event action sequence fixed.
This controlled comparison distinguishes event-associated regressions from failures already present without the change.
Across fourteen current VLA, hybrid, and world-action policies, events reduce success by 11.0–25.7 percentage points.
Event profiles reveal shared vulnerabilities to geometry and observation changes, while policy-family rankings interleave. Camera controls show that robustness reflects both competence under the changed conditions and the trajectory from which they are encountered; varying query cadence does not eliminate the gap. Together, the paired protocol and temporal diagnostics establish \maxbench as a reproducible testbed for diagnosing failures under mid-execution changes and measuring progress toward robot policies that remain effective as the world changes.

\end{abstract}

\section{Introduction}
Reliable robot control requires completing tasks even when the conditions that guided earlier actions no longer hold. A target may move during a reach, a camera may shift, or an obstacle may appear along the intended path. Simulation benchmarks provide controlled, repeatable settings for evaluating robot behavior~\citep{yu2020meta,james2020rlbench,mees2022calvin,mu2021maniskill,gu2023maniskill2}. Yet, robustness evaluations often configure external conditions before the first action and retain them throughout the rollout. This leaves a distinct question unresolved: \emph{How reliably can current robot policies complete a task when the world changes after they have begun to act?}

The timing of a change matters because it determines the history from which the policy must respond. A change introduced at reset tests whether a policy can operate under altered conditions from its first observation. A mid-task event requires the policy to continue from a state reached using earlier observations and actions. A camera shift changes the visual evidence available for control, while target relocation can invalidate a reach already in progress. Progress made before the change can also leave the robot closer to completing the task. For chunked policies, previously committed actions may continue to execute before the policy receives a post-event observation~\citep{black2025pi0,kim2025oft}. Moreover, no fixed action-chunk length is uniformly optimal across tasks~\citep{liang2026adaptive}. Task success therefore reflects both competence under the changed conditions and the state and action commitments inherited from earlier execution.\looseness-1

Existing benchmarks reveal failures under visual, geometric, linguistic, and semantic variations~\citep{fei2025liberoplus,zhou2025liberopro,wang2026liberox}, while DynamicVLA and DOMINO directly study moving-object manipulation and reactive control~\citep{xie2026dynamicvla,fang2026dynamicmanipulation}. These settings motivate a complementary evaluation: measuring the effect of a specific mid-execution event by comparing task completion with and without that event after the same observed and executed history. Independently sampled rollouts can diverge before the change, confounding its contribution with reset difficulty, policy randomness, and earlier actions. A controlled comparison must therefore hold the pre-event history fixed and measure whether task completion is preserved.\looseness-1

\begin{figure*}[t]
    \centering
    \includegraphics[width=\linewidth]{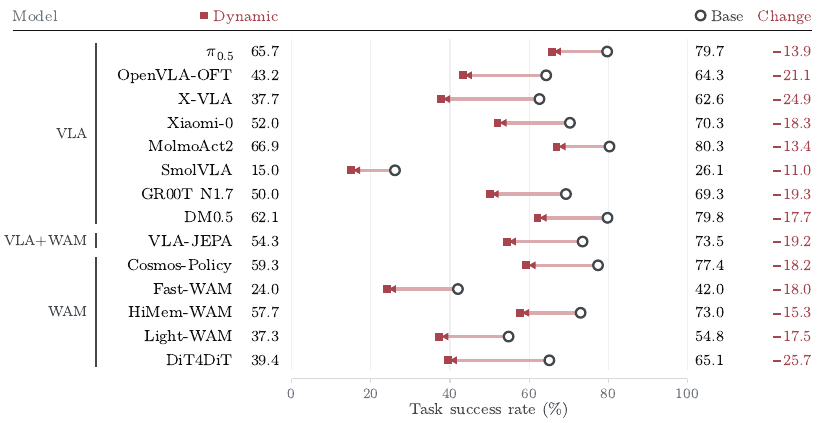}
    \caption{
    \textbf{One controlled mid-task change reduces success for every policy.}
    We evaluate each policy on the same 8,000 pairs using its released inference settings. Arrows connect Base (open circle) to Dynamic (filled square); columns report success rates (\%) and paired Dynamic minus Base changes in percentage points. Each Base/Dynamic pair shares the task, initial state, policy seed, and executed prefix, so the difference measures the effect of adding the event.}
    \label{fig:overall}
    \vspace{-0.3cm}
\end{figure*}

In this work, we introduce \textbf{\maxbench}, a benchmark for controlled evaluation of robot policies under mid-execution environmental changes. \maxbench comprises 8,000 rollout pairs, derived from LIBERO-Plus and LIBERO-PRO and balanced across eight event types. Each pair shares the task, initial simulator state, instruction, policy seed, and all actions executed before the event. This shared action sequence, the \emph{executed prefix}, establishes a common history for two continuations: \textbf{Base} proceeds without an added event, while \textbf{Dynamic} receives one event change and continues in the altered scene. For example, both rollouts approach the same target, which is then relocated only in Dynamic. Comparing their outcomes isolates the effect of introducing the event on task completion, holding the preceding execution fixed. Fig.~\ref{fig:overview} illustrates the changes, and Fig.~\ref{fig:overall} previews their aggregate effect.

We evaluate fourteen pretrained policies spanning vision-language-action (VLA) models, hybrid predictive, and world-action model (WAM) designs. Every policy loses 11.0--25.7 percentage points in success rate, and 20.8\%--56.1\% of episodes solved in Base fail after the event. MolmoAct2 and $\pi_{0.5}$ achieve the highest Dynamic success rates, but the remaining family rankings interleave: current WAMs are neither uniformly stronger nor uniformly weaker than VLAs. Relocation and camera/sensor changes produce the largest shared losses. On the same 800 paired cases, we vary how many actions policies execute before receiving a new observation and predicting their next actions. Changing this interval affects performance, but at every tested setting, success remains lower when an event occurs than when it does not. More frequent feedback therefore does not automatically eliminate the failures. Our contributions can be summarized as follows:
\begin{enumerate}[itemsep=0.5ex, parsep=0pt, topsep=-2pt, leftmargin=1cm]
    \item[\textbf{(1)}] \textbf{A paired benchmark for robustness during execution.} We introduce \maxbench, comprising 8,000 rollout pairs across diverse event types. Its Base/Dynamic protocol holds the executed action prefix fixed, isolating the effect of environmental changes introduced after action commitment.

 \item[\textbf{(2)}] \textbf{A systematic study of how online changes disrupt task success.}
    Across pretrained VLA, hybrid, and WAM policies, every paired success-rate gap is negative and statistically resolved. Paired outcome transitions quantify how often episodes that succeed in Base fail after the event.

 \item[\textbf{(3)}] \textbf{Diagnostics that connect event sensitivity with response timing.}
    We analyze event profiles and paired outcome transitions with trigger/response coverage, stale-action exposure, and query-cadence sweeps. These analyses reveal shared vulnerabilities to geometry and observation changes, interleaved VLA and WAM rankings, and persistent performance gaps across all tested feedback intervals.

\end{enumerate}

\section{Related Work}
\label{sec:related_work}
\textbf{Robot benchmarks and online changes.}
Simulation benchmarks support robot learning through controlled, repeatable evaluations of transfer and language-conditioned manipulation across diverse tasks~\citep{yu2020meta,james2020rlbench,mees2022calvin,gu2023maniskill2,nasiriany2024robocasa}. LIBERO and its Plus and PRO extensions evaluate transfer, OOD robustness, and memorization~\citep{liu2023libero,fei2025liberoplus,zhou2025liberopro}; related stress tests examine visual and embodiment variation~\citep{pumacay2024colosseum,wang2025vlatest,zhang2024vlabench,li24simpler}. ORION studies generalization from one human video to novel objects and varied scenes~\citep{zhu_vision_based_manipulation_auro_2026}. These evaluations usually set external conditions at reset. DynamicVLA and DOMINO study moving-object manipulation and reactive control~\citep{xie2026dynamicvla,fang2026dynamicmanipulation}. A matched comparison can distinguish the effect of a specific mid-task event from differences in the preceding execution. \maxbench provides this comparison by pairing no-event and event rollouts after an identical executed prefix across abrupt nuisance, observation, geometry, and path changes.

\textbf{Policies, prediction, and action commitment.}
VLA policies combine language and visual representations with action prediction~\citep{kim2024openvla,kim2025oft,black2025pi0,pi05,zheng2026xvla}. World-action and hybrid policies further connect prediction to control through video generation, latent actions, joint training, or memory~\citep{hu2025video,cen2025worldvla,bi2025motus,sun2026himemwam,li2026lightwam}. Chunked policies execute several predicted actions before observing and querying again, trading fewer policy calls for delayed feedback. Adaptive chunking adjusts this commitment across tasks~\citep{liang2026adaptive}. Success under conditions fixed at reset leaves open whether policies remain effective after changes during execution. \maxbench compares task completion after an external event across these policy designs and measures how query cadence changes outcomes. Its fixed event boundary supports controlled evaluation of future methods that cancel pending actions or query again after a change. Appendix~\ref{sec:extended_related_work} covers additional related work.

\section{The \maxbench{} Benchmark}

\subsection{Paired evaluation}
\maxbench measures the effect of a mid-task event by comparing two rollouts with the same \emph{executed prefix}, the sequence of actions performed before the event. For a fixed policy, each case specifies a task, instruction, initial state, policy seed, and event. We run each case in two conditions. Base retains the original source configuration without an additional mid-task event, including any variations inherited from LIBERO-Plus or LIBERO-PRO. Dynamic replays the Base actions up to trigger time $t_i$, then receives the event and continues in the changed environment.

Let $Y_i^0,Y_i^1\in\{0,1\}$ denote task completion in Base and Dynamic. Over $N$ pairs, the average success-rate change is $\widehat{\Delta}=\frac{1}{N}\sum_{i=1}^{N}(Y_i^1-Y_i^0)$. We report $100\widehat{\Delta}$ in percentage points. We retain every assigned case and its observed task outcome, including cases that do not reach the event trigger or obtain a post-event query. The four outcomes are preserved success $(1,1)$, change-associated success $(0,1)$, event-associated regression $(1,0)$, and persistent failure $(0,0)$.

The shared executed prefix is essential: independently sampled rollouts can approach the same task differently, whereas replay fixes the actions preceding the event. Beyond success rates, we report the fraction of all assigned cases that trigger the event and the fraction with a subsequent policy query. These measures distinguish reaching the change from obtaining post-event feedback. If a policy never reaches the trigger region, its task outcome still contributes to the success rate, while trigger coverage records that no event occurred. Figs.~\ref{fig:overview} and~\ref{fig:lineage} show pairing and case construction.

\vspace{-1.5mm}
\subsection{Construction and evaluation tracks}

\begin{figure*}[t]
    \centering
    \includegraphics[width=\linewidth]{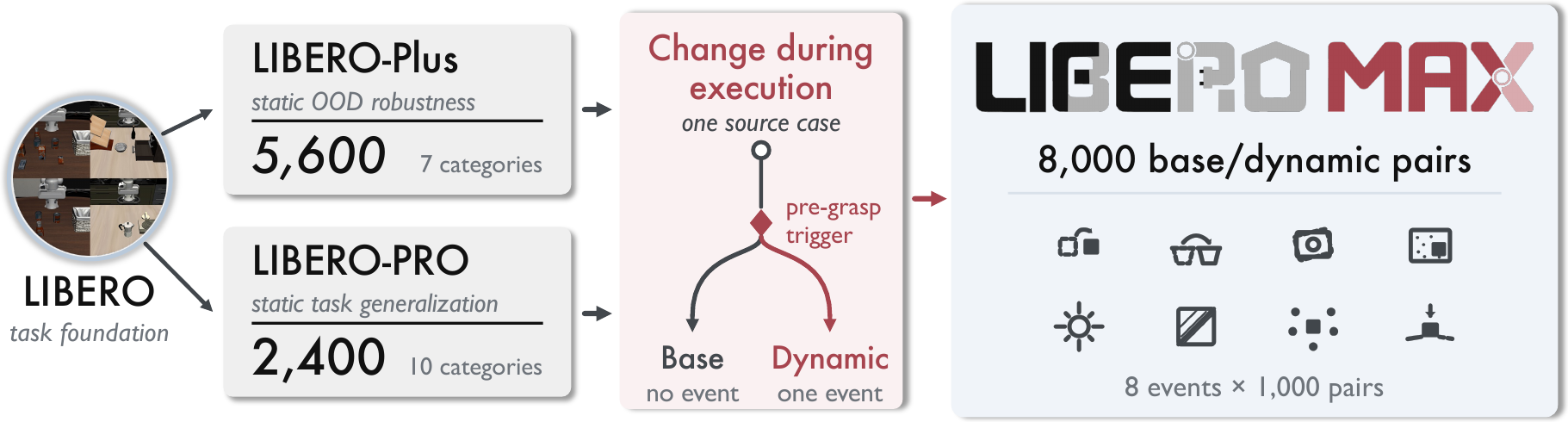}
    \caption{
    \textbf{From static source tasks to changes during execution.}
    LIBERO-Plus and LIBERO-PRO contribute 5,600 and 2,400 source cases. Each becomes a Base/Dynamic pair with the same actions before one mid-task event. Max contains 8,000 pairs, balanced at 1,000 per event.
    }
    \label{fig:lineage}
    \vspace{-3mm}
\end{figure*}

Max contains 5,600 cases from seven LIBERO-Plus categories and 2,400 from ten LIBERO-PRO categories, giving 16,000 Base/Dynamic rollouts per policy. Each of the eight event types has 1,000 pairs, split 700/300 between sources, with two fixed parameter variants. We select cases without using policy outcomes and replace invalid configurations within the same source category, event type, and parameter variant. Before evaluation, we check simulator validity, including support, workspace, visibility, and collision constraints for geometry changes, and retain only configurations that pass. Table~\ref{tab:benchmark_compare} compares evaluation scope, and Fig.~\ref{fig:lineage} shows the construction.

\begin{table*}[t]
\caption{\textbf{Benchmark scope and evaluation unit.} \cmark\ marks a directly evaluated property. LIBERO scores 2,000 task episodes, LIBERO-Plus 10,030 generated variants, LIBERO-PRO 10,000 task episodes, and \maxbench 8,000 Dynamic episodes with one no-event Base control per case. Only \maxbench applies an online change after an identical executed prefix.}
\label{tab:benchmark_compare}
\centering
\scriptsize
\setlength{\tabcolsep}{2.6pt}
\renewcommand{\arraystretch}{1.18}
\begin{tabularx}{\linewidth}{@{}l *{5}{Z} r@{}}
\toprule
\multicolumn{1}{@{}l}{\tableheader Benchmark} & \multicolumn{1}{c}{\tableheader Static} & \multicolumn{1}{c}{\tableheader Online} & \multicolumn{1}{c}{\tableheader Paired} & \multicolumn{1}{c}{\tableheader Exact} & \multicolumn{1}{c}{\tableheader Dynamic} & \multicolumn{1}{r@{}}{\tableheader Scored} \\
\multicolumn{1}{@{}l}{} & \multicolumn{1}{c}{\tableheader OOD} & \multicolumn{1}{c}{\tableheader change} & \multicolumn{1}{c}{\tableheader control} & \multicolumn{1}{c}{\tableheader prefix} & \multicolumn{1}{c}{\tableheader metrics} & \multicolumn{1}{r@{}}{\tableheader episodes} \\
\midrule
LIBERO & \xmark & \xmark & \xmark & \xmark & \xmark & \textbf{2,000} \\
LIBERO-Plus & \cmark & \xmark & \xmark & \xmark & \xmark & \textbf{10,030} \\
LIBERO-PRO & \cmark & \xmark & \xmark & \xmark & \xmark & \textbf{10,000} \\
\rowcolor{controlblue!8}
\textbf{\maxbench} & \cmark & \cmark & \cmark & \cmark & \cmark & \textbf{8,000} \\
\bottomrule
\end{tabularx}
\vspace{-3mm}
\end{table*}

For rapid iteration, \textbf{LIBERO-MAX Lite} fixes 800 pairs selected without policy outcomes, preserving Max's event and source proportions. It requires 1,600 rollouts per policy, one tenth of Max. Across fourteen policies, Base and Dynamic success rates and paired gaps stay within 2.4 percentage points of Max. We retain Max for detailed event and policy comparisons (Appendix~\ref{sec:max_lite}).

\vspace{-1.5mm}
\subsection{Events and temporal protocol}

The eight events are target and receptacle relocation, camera shift, illumination switch, sensor-noise onset, visual-theme switch, distractor burst, and obstacle insertion. Each is introduced once and is not reversed by the evaluator before the rollout ends; this is a \emph{persistent event}. For example, a camera shift retains the new viewpoint, while objects continue to obey the simulator's dynamics. An illumination switch leaves task geometry fixed; target relocation changes where the robot must reach, and obstacle insertion changes the approach region. These events require different responses, from continuing under altered observations to revising the target or path. Every event uses the deterministic \texttt{pre\_grasp\_proximity} trigger, which fires when the end effector first enters the designated entity's neighborhood during approach. Examples are shown in Appendix~\ref{sec:qualitative_examples}.

Base records policy queries and actions; Dynamic restores the same initial state and replays the exact prefix before applying the event. The evaluator checks prefix identity and records the trigger and first post-event query. A chunked policy predicts several actions at once, so it may execute actions queued before the event before seeing a new observation. We call their number \emph{stale-action exposure}.
All cases count toward success rates, including those without an event trigger or a post-event query. Protocol and validity details are in Appendix~\ref{sec:selection_audit}.

\vspace{-1mm}
\section{Results}
\vspace{-1.5mm}
\subsection{Evaluation protocol}
\label{sec:evaluation_protocol}
\vspace{-1.5mm}
We evaluate fourteen pretrained policies on the same 8,000 pairs: eight VLAs, VLA-JEPA with a predictive objective, and five WAMs. Each uses its released serving protocol, so scale, training, and action interfaces remain part of the policy comparison. Mimic-Video is evaluated separately on its supported partition. We report Base and Dynamic success rates, paired changes, conditional regression among Base successes, and event-family breakdowns. Dynamic success measures completion over all assigned cases; conditional regression measures the share of each policy's Base successes lost after the event. Paired changes use 95\% bootstrap intervals, and exact McNemar tests compare discordant outcomes. Cross-policy Dynamic contrasts use common case IDs with Holm correction. Model and evaluation details are in Appendix~\ref{sec:model_details}.

\vspace{-1.5mm}
\subsection{Online events break previously successful episodes}
\vspace{-1.5mm}

\begin{figure}[t]
    \centering
\includegraphics[width=0.95\linewidth]{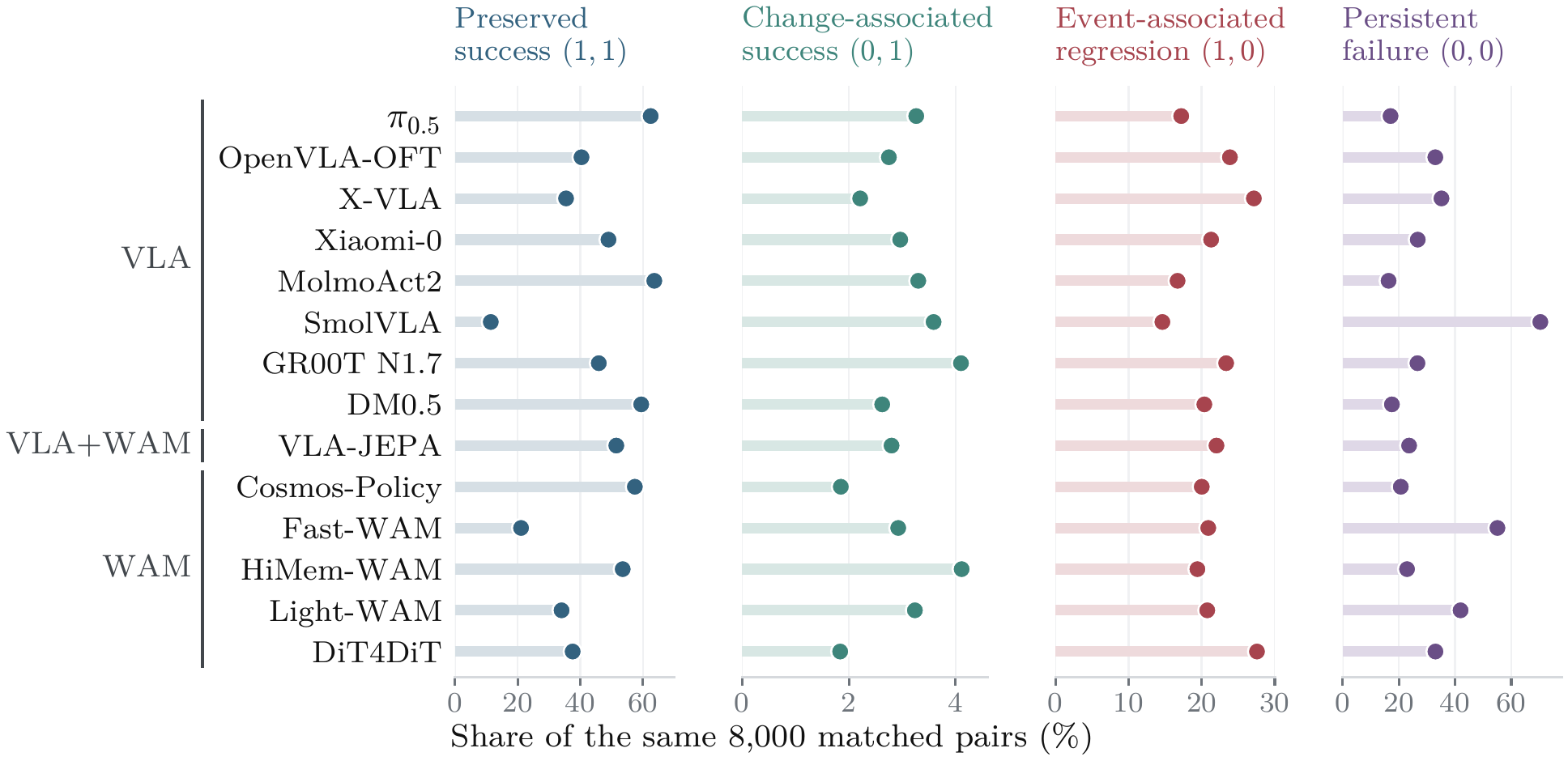}
    \caption{\textbf{Paired outcomes separate post-change regression from low Base competence.} Panels partition all 8,000 pairs into the four binary outcomes. Points show episode shares; independently scaled axes retain visibility of rare change-associated successes $(0,1)$. The $(1,0)$ group counts cases solved in Base but failed in Dynamic; $(0,1)$ counts the reverse outcome.}
    \label{fig:outcomes}
    \vspace{-2mm}
\end{figure}

Every policy is less successful after the event (Table~\ref{tab:main}, Fig.~\ref{fig:overall}). Even policies with strong Base performance lose success: MolmoAct2 falls from 80.3\% to 66.9\%, and $\pi_{0.5}$ from 79.7\% to 65.7\%.
Fig.~\ref{fig:outcomes} separates cases that fail only after the event from those that already fail in Base. SmolVLA has the smallest aggregate gap, but only 26.1\% Base success; the event breaks 56.1\% of those successes. A few cases succeed only in Dynamic $(0,1)$, partly offsetting these losses in the average. Conditional regression counts the share of Base successes lost after the event, making these failures visible even when the average gap is small. Full paired diagnostics are in Appendix~\ref{sec:paired_diagnostics}.

MolmoAct2 has the highest Dynamic success rate, yet still loses many cases it solves in Base. It preserves 5,091 Base successes but loses another 1,336 after the event; its 264 change-associated successes partially offset those losses in the average. Thus its 13.4-point net decline corresponds to the loss of 20.8\% of previously solved episodes. A higher Dynamic rate can come from preserving more Base successes, solving different cases, or both; paired outcomes distinguish these possibilities. Trigger and response coverage range from 87.7\% to 96.6\% across policies. Most cases therefore reach the event and receive a new observation, yet many still fail. Cases that never reach the trigger still count toward the success rate.

\begin{table*}[t]
\caption{\textbf{Every policy loses success after a controlled online change.} All fourteen rows contain the same 8,000 Base/Dynamic pairs. Overall $\Delta$ and the four event-family columns report Dynamic minus Base success in percentage points. Released serving protocols differ across rows, so the table compares policies rather than isolating an architecture-family effect.}
\label{tab:main}
\centering
\scriptsize
\setlength{\tabcolsep}{2.4pt}
\renewcommand{\arraystretch}{1.18}
\begin{tabularx}{\linewidth}{@{}llccZZZcZ@{}}
\toprule
\multirow[c]{2}{*}[-0.55ex]{\tableheader Family} &
\multirow[c]{2}{*}[-0.55ex]{\tableheader Model} &
\multicolumn{3}{c}{\tableheader Overall success rate (\%)} &
\multicolumn{4}{c}{\tableheader Change family (paired $\Delta$, pp)} \\
\cmidrule(lr){3-5}\cmidrule(l){6-9}
& & \tableheader Base SR $\uparrow$ & \tableheader Dynamic SR $\uparrow$ & \multicolumn{1}{c}{\tableheader $\Delta$} &
\multicolumn{1}{c}{\tableheader Obs.} & \multicolumn{1}{c}{\tableheader Geom.} & \tableheader App.\,+ clutter & \multicolumn{1}{c}{\tableheader Path} \\
\midrule
\rowgray & $\pi_{0.5}$ & 79.7 & 65.7 & $-13.9$ & $-23.1$ & $-24.2$ & $-4.2$ & $-4.2$ \\
& OpenVLA-OFT & 64.3 & 43.2 & $-21.1$ & $-32.0$ & $-33.6$ & $-10.3$ & $-6.8$ \\
\rowgray & X-VLA & 62.6 & 37.7 & $-24.9$ & $-43.0$ & $-52.4$ & $-1.2$ & $-5.1$ \\
& Xiaomi-Robotics-0 & 70.3 & 52.0 & $-18.3$ & $-25.2$ & $-37.7$ & $-5.7$ & $-3.9$ \\
\rowgray & MolmoAct2 & 80.3 & 66.9 & $-13.4$ & $-19.3$ & $-26.6$ & $-4.6$ & $-1.8$ \\
& SmolVLA & 26.1 & 15.0 & $-11.0$ & $-16.6$ & $-21.2$ & $-3.3$ & $-2.7$ \\
\rowgray & GR00T N1.7 & 69.3 & 50.0 & $-19.3$ & $-24.6$ & $-45.0$ & $-2.2$ & $-8.3$ \\
\multirow{-8}{*}{VLA} & DM0.5 & 79.8 & 62.1 & $-17.7$ & $-29.0$ & $-31.3$ & $-5.2$ & $-5.7$ \\
\midrule
VLA+WAM & VLA-JEPA & 73.5 & 54.3 & $-19.2$ & $-30.3$ & $-33.7$ & $-7.3$ & $-4.0$ \\
\midrule
\rowgray & Cosmos-Policy & 77.4 & 59.3 & $-18.2$ & $-25.2$ & $-35.8$ & $-6.3$ & $-4.6$ \\
& Fast-WAM & 42.0 & 24.0 & $-18.0$ & $-35.6$ & $-27.5$ & $-3.9$ & $-6.1$ \\
\rowgray & HiMem-WAM & 73.0 & 57.7 & $-15.3$ & $-13.3$ & $-39.8$ & $-4.3$ & $-3.3$ \\
& Light-WAM & 54.8 & 37.3 & $-17.5$ & $-21.7$ & $-39.9$ & $-5.0$ & $-2.0$ \\
\rowgray \multirow{-5}{*}{WAM} & DiT4DiT & 65.1 & 39.4 & $-25.7$ & $-49.7$ & $-26.6$ & $-16.2$ & $-4.5$ \\
\bottomrule
\end{tabularx}
\end{table*}

\rqanswer{Takeaway 1: Online events break policies that otherwise succeed.}{Across policies, events reduce success by 11.0 to 25.7 percentage points and turn 20.8\% to 56.1\% of Base successes into failures. SmolVLA loses the fewest percentage points overall but the largest share of its Base successes because its Base success rate is low.}

\subsection{Policy families have no consistent ordering}

Dynamic rankings and paired losses interleave across policy families. MolmoAct2 and $\pi_{0.5}$ lead Dynamic success at 66.9\% and 65.7\%, followed by DM0.5 (62.1\%), Cosmos-Policy (59.3\%), and HiMem-WAM (57.7\%). \looseness-1

\rqanswer{Takeaway 2: Current WAMs do not consistently outperform or underperform VLAs.}{The three highest Dynamic success rates belong to VLAs, followed by two WAMs. The remaining rankings and the size of the losses show no consistent VLA--WAM ordering. These policies also differ in training, scale, and inference settings, so the comparison cannot isolate the effect of architecture.}

\subsection{Geometry and observation changes cause the largest losses}

\begin{figure*}[t]
    \centering
    \includegraphics[width=\linewidth]{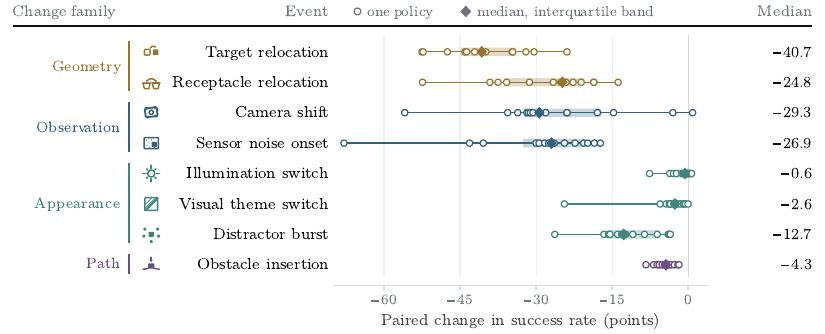}
    \caption{\textbf{Geometry and observation changes produce the largest shared losses.} Rows group eight events into four families. Open circles show fourteen policies per 1,000-pair event; diamonds, shading, and thin segments mark the median, interquartile range, and full range.}
    \label{fig:events}
    \vspace{-3mm}
\end{figure*}

Event profiles reveal where these robustness losses are concentrated. Fig.~\ref{fig:events} shows a shared event ordering, with mean pairwise Spearman correlation $0.80$ across policies. Geometry losses span 21.2--52.4 points and observation losses 13.3--49.7, exceeding most appearance-and-clutter and path losses. Individual profiles differ: Fast-WAM falls from 43.4\% Base success to 0.3\% under sensor noise, whereas X-VLA is most sensitive to camera shift. Illumination and theme changes generally preserve more success than relocating the target or destination, but the exceptions matter: sensor noise dominates Fast-WAM and DiT4DiT, while HiMem-WAM is comparatively camera robust. Thus event profiles help identify which changed conditions a policy handles poorly, beyond a single family ranking. Full event-level results are in Appendix~\ref{sec:event_diagnostics}.

\textbf{Source and task variation.}
The gap persists when source categories receive equal weight: category-macro losses range from 10.7 to 23.4 points across all fourteen policies, and every policy loses success in each of the four LIBERO task suites. The hardest suite nevertheless depends on the policy. MolmoAct2 loses 16.0 points on LIBERO-10 versus 9.6 on LIBERO-Object, whereas Light-WAM loses 25.3 points on LIBERO-Spatial versus 12.1 on LIBERO-10. Breaking down results by event and task shows where each policy struggles, even when its overall success rate is high. Full breakdowns are in Appendix~\ref{sec:factorized_diagnostics}.

Base success also matters when comparing source categories. PRO-derived cases have lower Dynamic success but also lower Base success and smaller paired losses than Plus-derived cases. SmolVLA succeeds on only 0.4\% of PRO position cases in Base, so its success rate can barely fall further. Base success and conditional regression help distinguish cases a policy already struggles with from cases it fails only after the event (Appendix~\ref{sec:source_diagnostics}).

\textbf{Event parameters expose further sensitivity.}
Across policies, the largest within-event difference in Dynamic success between the two fixed parameter variants ranges from 12.2 to 36.4 points; the difference reaches 35.6 points for $\pi_{0.5}$ under camera shift and 36.4 points for HiMem-WAM under receptacle relocation. Success can therefore vary substantially within the same event type. Because the variants contain different source instances, the comparison does not isolate the effect of changing the parameter. Each policy is evaluated on the same cases from both parameter variants. Reporting the groups separately shows whether an improvement extends across the predefined configurations or is concentrated in one of them. Medium/high severity comparisons also combine different event types and should be read as descriptions of those groups (Appendix~\ref{sec:factorized_diagnostics}).

\rqanswer{Takeaway 3: Target relocation causes the largest loss for 11 of 14 policies.}{On \maxbench, target relocation is the most damaging tested event for eleven policies. X-VLA instead loses the most under camera shift, while Fast-WAM and DiT4DiT lose the most under sensor noise. Illumination and visual-theme changes are generally milder.}

\subsection{Camera controls reveal the roles of viewpoint and execution history}

\begin{wraptable}{R}{0.57\textwidth}
    \vspace{-3mm}
    \caption{\textbf{Changed views can impair success from reset.} SR (\%) on all 1,000 cases per policy. Bold: estimates; italic below: 95\% source-stratified bootstrap CIs.}
    \label{tab:camera_timing_success}
    \centering
    \scriptsize
    \setlength{\tabcolsep}{1.7pt}
    \renewcommand{\arraystretch}{1.04}
    \begin{tabular*}{\linewidth}{@{\extracolsep{\fill}}lccc@{}}
    \toprule
    \tableheader Policy & \tableheader Base & \tableheader Reset & \tableheader Mid-task \\
    \midrule
    \multirow[c]{2}{*}{X-VLA} & \textbf{68.2} & \textbf{1.3} & \textbf{12.4} \\
    & \textit{[65.4, 71.0]} & \textit{[0.6, 2.0]} & \textit{[10.4, 14.5]} \\
    \addlinespace[2pt]
    \multirow[c]{2}{*}{$\pi_{0.5}$} & \textbf{79.5} & \textbf{51.3} & \textbf{55.3} \\
    & \textit{[77.1, 81.9]} & \textit{[48.2, 54.4]} & \textit{[52.2, 58.4]} \\
    \addlinespace[2pt]
    \multirow[c]{2}{*}{HiMem-WAM} & \textbf{72.5} & \textbf{68.8} & \textbf{72.6} \\
    & \textit{[69.7, 75.2]} & \textit{[65.9, 71.6]} & \textit{[69.8, 75.3]} \\
    \bottomrule
    \end{tabular*}
    \vspace{-2mm}
\end{wraptable}

The large losses under observation changes raise a question: how much of the camera-shift difficulty is already present when a policy starts in the changed view? We run an independent control experiment on all 1,000 camera cases for X-VLA, $\pi_{0.5}$, and HiMem-WAM (700 Plus and 300 PRO cases per policy; 9,000 rollouts). Base retains the original camera. Reset applies the camera change before the first policy input. Mid-task follows the Dynamic protocol, applying the same change after replaying the Base actions up to the trigger. Across these three conditions, the cases, initial states, checkpoints, seeds, and success criteria remain fixed.

The changed view can be difficult even from the start (Table~\ref{tab:camera_timing_success}). X-VLA falls from 68.2\% Base success to 1.3\% at Reset and reaches 12.4\% at Mid-task. Mid-task exceeds Reset by 11.1, 4.0, and 3.8 points for X-VLA, $\pi_{0.5}$, and HiMem-WAM, respectively. Starting in the original view can therefore help a policy make progress before the change. Reset and Mid-task differ in visited states, time spent in the changed view, and remaining task budget, whereas Base and Mid-task share the same executed prefix.\looseness-1

\textbf{Consistency across sources and task groups.}
All three policies have higher Mid-task than Reset success in both the Plus and PRO groups. When we combine the sources, the advantage remains statistically supported after resampling the 40 shared reference tasks, source-specific task groups, or initial-configuration groups. The smaller PRO group gives less precise estimates: its results alone do not establish an advantage for $\pi_{0.5}$ or HiMem-WAM (Appendix~\ref{sec:camera_timing_control}).

\textbf{Which trajectories lose success?}
Similar success rates can hide different outcomes on individual cases. HiMem-WAM's nearly unchanged Base/Mid success (72.5\%/72.6\%) combines 66 cases that lose success with 67 that gain it. Across the three policies, 84 model--case instances succeed in both Base and Reset but fail in Mid-task. All receive a new policy query, and 13 execute no old queued actions after the event. Thus some mid-task failures occur even when the policy can solve the case from the changed view at reset, and stale actions alone cannot explain all of them.

We also check whether the robot reaches the trigger's 18\,cm neighborhood. Among the 106 X-VLA cases solved in Base and Mid-task but failed in Reset, 74 Reset trajectories never enter this region; the corresponding count is only 2 of 59 for HiMem-WAM. In these cases, X-VLA often struggles to approach the target from the changed view, whereas HiMem-WAM usually fails after reaching the neighborhood. Entering this region marks approach, not a completed grasp.\looseness-1

\textbf{The timing advantage persists at milder camera shifts.}
On the same 1,000 cases, reducing camera magnitude to 25\% raises X-VLA Reset success to 36.0\% and Mid-task success to 51.7\%, compared with 68.2\% in Base. Both improve but remain below Base. Mid-task still outperforms Reset when Reset success is well above its full-strength near-zero level. The timing gap peaks at 50\% strength rather than increasing monotonically with magnitude. At the milder settings, most Reset failures among cases solved in Base and Mid-task occur after reaching the approach threshold. These controls show that both the viewpoint and the preceding trajectory matter for completing the task (Appendix~\ref{sec:camera_timing_control}).

\subsection{Query cadence changes performance without closing the gap}

\begin{figure*}[t]
    \centering
    \includegraphics[width=\linewidth]{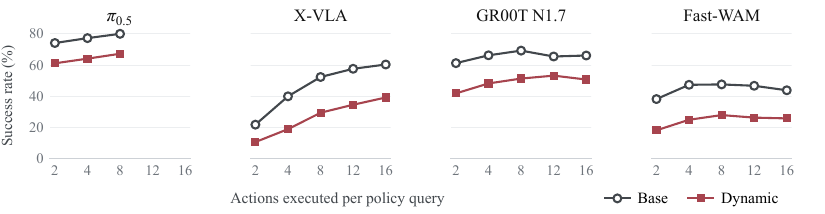}
    \caption{\textbf{Dynamic remains below Base at every valid query cadence.} Panels use the same 800 matched pairs per policy. Open circles are Base and filled squares Dynamic; $Q$ is the executed actions per query. Only $Q\leq H$ is valid, so $\pi_{0.5}$ stops at $Q=8$.}
    \label{fig:action_cadence}
    \vspace{-2.5mm}
\end{figure*}

We next examine whether changing the frequency of policy feedback can recover Dynamic success. The action horizon $H$ is the number of actions returned by a policy call; the query interval $Q$ is the number executed before a new observation and call. We vary $Q$ on the fixed 800-pair Lite pool (Fig.~\ref{fig:action_cadence}). X-VLA, GR00T N1.7, and Fast-WAM use $Q\in\{2,4,8,12,16\}$; $\pi_{0.5}$ uses $\{2,4,8\}$ because its $H=10$ horizon excludes longer prefixes. Fast-WAM fixes $H=32$, while X-VLA and GR00T N1.7 use $H=Q$. Thus the sweep measures each checkpoint's practical serving protocol.

X-VLA and $\pi_{0.5}$ improve as $Q$ increases over their valid ranges; GR00T N1.7 and Fast-WAM peak in Dynamic success at $Q=12$ and $Q=8$. Across all 18 settings, Dynamic remains 11.1--23.0 points below Base. This policy-specific pattern agrees with evidence that the best fixed chunk length varies across tasks~\citep{liang2026adaptive}. In the primary evaluation, policies often continue executing queued actions after the event. The median is two actions for $\pi_{0.5}$ and seven for Cosmos-Policy and Fast-WAM. These actions delay the next observation, but no tested query cadence closes the gap (Appendix~\ref{sec:cadence_sweep}).

For X-VLA, Dynamic success rises from 10.8\% at $Q=2$ to 39.2\% at $Q=16$, although the Base--Dynamic gap grows. A smaller gap alone therefore does not imply better task completion.

\rqanswer{Takeaway 4: More frequent queries alone do not close the response gap.}{For the six policies with execution traces, the average number of queued actions executed after an event before the next observation ranges from 2.02 to 7.47. On the fixed 800-pair sweep, Dynamic success stays below Base at every tested query interval. The best $Q$ depends on the policy, and changing it improves some results without eliminating the loss.}

\noindent\begin{minipage}{\textwidth}
\begin{wraptable}{r}{0.5\textwidth}
    \caption{\textbf{Restoration recovers part of the sensor-noise loss.} X-VLA on 300 cases. Bold: SR (\%) and Dynamic gain (pp) over Native; italic below: paired 95\% bootstrap CIs for gains.}
    \label{tab:sensor_recovery}
    \centering
    \scriptsize
    \setlength{\tabcolsep}{3pt}
    \renewcommand{\arraystretch}{1.04}
    \begin{tabular*}{\linewidth}{@{\extracolsep{\fill}}lccc@{}}
        \toprule
        \tableheader Processing & \tableheader Base & \tableheader Dynamic & \tableheader Gain \\
        \midrule
        Native & \textbf{66.7} & \textbf{40.7} & --- \\
        \addlinespace[2pt]
        \multirow[c]{2}{*}{Always-on} & \multirow[c]{2}{*}{\textbf{68.3}} & \multirow[c]{2}{*}{\textbf{47.3}} & \textbf{+6.7} \\
        & & & \textit{[2.3, 11.0]} \\
        \addlinespace[2pt]
        \multirow[c]{2}{*}{Quality-gated} & \multirow[c]{2}{*}{\textbf{67.0}} & \multirow[c]{2}{*}{\textbf{47.7}} & \textbf{+7.0} \\
        & & & \textit{[3.0, 11.3]} \\
        \bottomrule
    \end{tabular*}
    \vspace{-2mm}
\end{wraptable}

\textbf{A candidate response to sensor corruption.}
We test whether restoring corrupted images can reduce the sensor-noise loss on 300 fixed cases. Before querying X-VLA, a quality gate checks the images and applies inpainting and denoising when needed; the policy itself is not retrained. The gate uses only the current images and receives neither an event flag nor a clean reference image. Table~\ref{tab:sensor_recovery} compares native observations, always-on restoration, and quality-gated restoration. The gated response raises Dynamic success from 40.7\% to 47.7\%, while Base changes from 66.7\% to 67.0\%. The Base/Dynamic gap therefore narrows from 26.0 to 19.3 points, a paired reduction of 6.7 points (95\% CI: [2.3, 11.3]). At the case level, gated restoration converts 33 native Dynamic failures to successes while losing 12 native successes. Always-on restoration also improves Dynamic success, and the data do not establish an advantage for gating over always-on processing. The benchmark can thus measure whether a targeted intervention reduces the loss after a change. Experimental details are in Appendix~\ref{sec:sensor_recovery}.
\end{minipage}

\section{Conclusion}

\maxbench measures whether robot policies can complete tasks after their environment changes during execution. Its paired Base/Dynamic protocol holds the task, initial state, policy seed, and executed action prefix fixed, allowing direct comparison with a no-event rollout. Across eight event types and fourteen current policies, online changes cause substantial failures, with the largest shared losses under geometry and observation changes. Camera controls show that the changed viewpoint and preceding trajectory both affect success. Paired outcomes reveal which previously successful cases fail, while execution traces show when the policy receives new observations. No policy family consistently outperforms the others, and changing query cadence leaves substantial losses at every tested setting. A simple image-restoration module recovers part of the sensor-noise loss. \maxbench can therefore support both identifying the changes that disrupt a policy and testing methods that help it continue the task.

\clearpage

\bibliographystyle{apalike}
\bibliography{iclr2027_conference}

\clearpage
\appendix
\makeatletter
\setlength{\@fptop}{0pt}
\makeatother
\begingroup
\hypersetup{linkcolor=MaxAccent}
\color{MaxAccent}
\begin{center}
{\sffamily\bfseries\fontsize{18}{21}\selectfont Appendix\par}
\end{center}
\phantomsection
\pdfbookmark[0]{Appendix}{appendix.contents}

\fontsize{10}{12}\selectfont
\setlength{\tabcolsep}{4pt}
\renewcommand{\arraystretch}{1.16}
\newcommand{\contentssection}[1]{{\fontsize{11}{13}\selectfont\bfseries #1}}
\noindent\begin{tabularx}{\linewidth}{@{}l >{\raggedright\arraybackslash}X r@{}}
\contentssection{\hyperref[sec:extended_related_work]{\ref*{sec:extended_related_work}}} & \contentssection{\hyperref[sec:extended_related_work]{\nameref*{sec:extended_related_work}}} & \hyperref[sec:extended_related_work]{\pageref*{sec:extended_related_work}} \\
\hspace*{0.8em}\hyperref[sec:related_benchmarks]{\ref*{sec:related_benchmarks}} & \hspace*{0.8em}\hyperref[sec:related_benchmarks]{\nameref*{sec:related_benchmarks}} & \hyperref[sec:related_benchmarks]{\pageref*{sec:related_benchmarks}} \\
\hspace*{0.8em}\hyperref[sec:related_objects]{\ref*{sec:related_objects}} & \hspace*{0.8em}\hyperref[sec:related_objects]{\nameref*{sec:related_objects}} & \hyperref[sec:related_objects]{\pageref*{sec:related_objects}} \\
\hspace*{0.8em}\hyperref[sec:related_vla]{\ref*{sec:related_vla}} & \hspace*{0.8em}\hyperref[sec:related_vla]{\nameref*{sec:related_vla}} & \hyperref[sec:related_vla]{\pageref*{sec:related_vla}} \\
\hspace*{0.8em}\hyperref[sec:related_wam]{\ref*{sec:related_wam}} & \hspace*{0.8em}\hyperref[sec:related_wam]{\nameref*{sec:related_wam}} & \hyperref[sec:related_wam]{\pageref*{sec:related_wam}} \\
\addlinespace[6pt]
\contentssection{\hyperref[sec:selection_audit]{\ref*{sec:selection_audit}}} & \contentssection{\hyperref[sec:selection_audit]{\nameref*{sec:selection_audit}}} & \hyperref[sec:selection_audit]{\pageref*{sec:selection_audit}} \\
\hspace*{0.8em}\hyperref[sec:benchmark_details]{\ref*{sec:benchmark_details}} & \hspace*{0.8em}\hyperref[sec:benchmark_details]{\nameref*{sec:benchmark_details}} & \hyperref[sec:benchmark_details]{\pageref*{sec:benchmark_details}} \\
\hspace*{0.8em}\hyperref[sec:case_construction]{\ref*{sec:case_construction}} & \hspace*{0.8em}\hyperref[sec:case_construction]{\nameref*{sec:case_construction}} & \hyperref[sec:case_construction]{\pageref*{sec:case_construction}} \\
\hspace*{0.8em}\hyperref[sec:event_protocol]{\ref*{sec:event_protocol}} & \hspace*{0.8em}\hyperref[sec:event_protocol]{\nameref*{sec:event_protocol}} & \hyperref[sec:event_protocol]{\pageref*{sec:event_protocol}} \\
\hspace*{0.8em}\hyperref[sec:validity_checks]{\ref*{sec:validity_checks}} & \hspace*{0.8em}\hyperref[sec:validity_checks]{\nameref*{sec:validity_checks}} & \hyperref[sec:validity_checks]{\pageref*{sec:validity_checks}} \\
\hspace*{0.8em}\hyperref[sec:qualitative_examples]{\ref*{sec:qualitative_examples}} & \hspace*{0.8em}\hyperref[sec:qualitative_examples]{\nameref*{sec:qualitative_examples}} & \hyperref[sec:qualitative_examples]{\pageref*{sec:qualitative_examples}} \\
\addlinespace[6pt]
\contentssection{\hyperref[sec:max_lite]{\ref*{sec:max_lite}}} & \contentssection{\hyperref[sec:max_lite]{\nameref*{sec:max_lite}}} & \hyperref[sec:max_lite]{\pageref*{sec:max_lite}} \\
\hspace*{0.8em}\hyperref[sec:lite_selection]{\ref*{sec:lite_selection}} & \hspace*{0.8em}\hyperref[sec:lite_selection]{\nameref*{sec:lite_selection}} & \hyperref[sec:lite_selection]{\pageref*{sec:lite_selection}} \\
\hspace*{0.8em}\hyperref[sec:lite_validation]{\ref*{sec:lite_validation}} & \hspace*{0.8em}\hyperref[sec:lite_validation]{\nameref*{sec:lite_validation}} & \hyperref[sec:lite_validation]{\pageref*{sec:lite_validation}} \\
\addlinespace[6pt]
\contentssection{\hyperref[sec:model_details]{\ref*{sec:model_details}}} & \contentssection{\hyperref[sec:model_details]{\nameref*{sec:model_details}}} & \hyperref[sec:model_details]{\pageref*{sec:model_details}} \\
\hspace*{0.8em}\hyperref[sec:protocol_details]{\ref*{sec:protocol_details}} & \hspace*{0.8em}\hyperref[sec:protocol_details]{\nameref*{sec:protocol_details}} & \hyperref[sec:protocol_details]{\pageref*{sec:protocol_details}} \\
\hspace*{0.8em}\hyperref[sec:policy_checkpoints]{\ref*{sec:policy_checkpoints}} & \hspace*{0.8em}\hyperref[sec:policy_checkpoints]{\nameref*{sec:policy_checkpoints}} & \hyperref[sec:policy_checkpoints]{\pageref*{sec:policy_checkpoints}} \\
\hspace*{0.8em}\hyperref[sec:policy_execution]{\ref*{sec:policy_execution}} & \hspace*{0.8em}\hyperref[sec:policy_execution]{\nameref*{sec:policy_execution}} & \hyperref[sec:policy_execution]{\pageref*{sec:policy_execution}} \\
\hspace*{0.8em}\hyperref[sec:mimic_partial]{\ref*{sec:mimic_partial}} & \hspace*{0.8em}\hyperref[sec:mimic_partial]{\nameref*{sec:mimic_partial}} & \hyperref[sec:mimic_partial]{\pageref*{sec:mimic_partial}} \\
\addlinespace[6pt]
\contentssection{\hyperref[sec:detailed_results]{\ref*{sec:detailed_results}}} & \contentssection{\hyperref[sec:detailed_results]{\nameref*{sec:detailed_results}}} & \hyperref[sec:detailed_results]{\pageref*{sec:detailed_results}} \\
\hspace*{0.8em}\hyperref[sec:paired_diagnostics]{\ref*{sec:paired_diagnostics}} & \hspace*{0.8em}\hyperref[sec:paired_diagnostics]{\nameref*{sec:paired_diagnostics}} & \hyperref[sec:paired_diagnostics]{\pageref*{sec:paired_diagnostics}} \\
\hspace*{0.8em}\hyperref[sec:event_diagnostics]{\ref*{sec:event_diagnostics}} & \hspace*{0.8em}\hyperref[sec:event_diagnostics]{\nameref*{sec:event_diagnostics}} & \hyperref[sec:event_diagnostics]{\pageref*{sec:event_diagnostics}} \\
\hspace*{0.8em}\hyperref[sec:source_diagnostics]{\ref*{sec:source_diagnostics}} & \hspace*{0.8em}\hyperref[sec:source_diagnostics]{\nameref*{sec:source_diagnostics}} & \hyperref[sec:source_diagnostics]{\pageref*{sec:source_diagnostics}} \\
\hspace*{0.8em}\hyperref[sec:factorized_diagnostics]{\ref*{sec:factorized_diagnostics}} & \hspace*{0.8em}\hyperref[sec:factorized_diagnostics]{\nameref*{sec:factorized_diagnostics}} & \hyperref[sec:factorized_diagnostics]{\pageref*{sec:factorized_diagnostics}} \\
\addlinespace[6pt]
\contentssection{\hyperref[sec:camera_timing_control]{\ref*{sec:camera_timing_control}}} & \contentssection{\hyperref[sec:camera_timing_control]{\nameref*{sec:camera_timing_control}}} & \hyperref[sec:camera_timing_control]{\pageref*{sec:camera_timing_control}} \\
\hspace*{0.8em}\hyperref[sec:camera_design]{\ref*{sec:camera_design}} & \hspace*{0.8em}\hyperref[sec:camera_design]{\nameref*{sec:camera_design}} & \hyperref[sec:camera_design]{\pageref*{sec:camera_design}} \\
\hspace*{0.8em}\hyperref[sec:camera_effects]{\ref*{sec:camera_effects}} & \hspace*{0.8em}\hyperref[sec:camera_effects]{\nameref*{sec:camera_effects}} & \hyperref[sec:camera_effects]{\pageref*{sec:camera_effects}} \\
\hspace*{0.8em}\hyperref[sec:camera_trajectories]{\ref*{sec:camera_trajectories}} & \hspace*{0.8em}\hyperref[sec:camera_trajectories]{\nameref*{sec:camera_trajectories}} & \hyperref[sec:camera_trajectories]{\pageref*{sec:camera_trajectories}} \\
\hspace*{0.8em}\hyperref[sec:camera_magnitude]{\ref*{sec:camera_magnitude}} & \hspace*{0.8em}\hyperref[sec:camera_magnitude]{\nameref*{sec:camera_magnitude}} & \hyperref[sec:camera_magnitude]{\pageref*{sec:camera_magnitude}} \\
\addlinespace[6pt]
\contentssection{\hyperref[sec:cadence_sweep]{\ref*{sec:cadence_sweep}}} & \contentssection{\hyperref[sec:cadence_sweep]{\nameref*{sec:cadence_sweep}}} & \hyperref[sec:cadence_sweep]{\pageref*{sec:cadence_sweep}} \\
\hspace*{0.8em}\hyperref[sec:cadence_design]{\ref*{sec:cadence_design}} & \hspace*{0.8em}\hyperref[sec:cadence_design]{\nameref*{sec:cadence_design}} & \hyperref[sec:cadence_design]{\pageref*{sec:cadence_design}} \\
\hspace*{0.8em}\hyperref[sec:stale_exposure]{\ref*{sec:stale_exposure}} & \hspace*{0.8em}\hyperref[sec:stale_exposure]{\nameref*{sec:stale_exposure}} & \hyperref[sec:stale_exposure]{\pageref*{sec:stale_exposure}} \\
\addlinespace[6pt]
\contentssection{\hyperref[sec:sensor_recovery]{\ref*{sec:sensor_recovery}}} & \contentssection{\hyperref[sec:sensor_recovery]{\nameref*{sec:sensor_recovery}}} & \hyperref[sec:sensor_recovery]{\pageref*{sec:sensor_recovery}} \\
\hspace*{0.8em}\hyperref[sec:recovery_design]{\ref*{sec:recovery_design}} & \hspace*{0.8em}\hyperref[sec:recovery_design]{\nameref*{sec:recovery_design}} & \hyperref[sec:recovery_design]{\pageref*{sec:recovery_design}} \\
\hspace*{0.8em}\hyperref[sec:recovery_effects]{\ref*{sec:recovery_effects}} & \hspace*{0.8em}\hyperref[sec:recovery_effects]{\nameref*{sec:recovery_effects}} & \hyperref[sec:recovery_effects]{\pageref*{sec:recovery_effects}} \\
\hspace*{0.8em}\hyperref[sec:recovery_cost]{\ref*{sec:recovery_cost}} & \hspace*{0.8em}\hyperref[sec:recovery_cost]{\nameref*{sec:recovery_cost}} & \hyperref[sec:recovery_cost]{\pageref*{sec:recovery_cost}} \\
\addlinespace[6pt]
\contentssection{\hyperref[sec:scope_limitations]{\ref*{sec:scope_limitations}}} & \contentssection{\hyperref[sec:scope_limitations]{\nameref*{sec:scope_limitations}}} & \hyperref[sec:scope_limitations]{\pageref*{sec:scope_limitations}} \\
\hspace*{0.8em}\hyperref[sec:discussion_interpretation]{\ref*{sec:discussion_interpretation}} & \hspace*{0.8em}\hyperref[sec:discussion_interpretation]{\nameref*{sec:discussion_interpretation}} & \hyperref[sec:discussion_interpretation]{\pageref*{sec:discussion_interpretation}} \\
\hspace*{0.8em}\hyperref[sec:discussion_limits]{\ref*{sec:discussion_limits}} & \hspace*{0.8em}\hyperref[sec:discussion_limits]{\nameref*{sec:discussion_limits}} & \hyperref[sec:discussion_limits]{\pageref*{sec:discussion_limits}} \\
\hspace*{0.8em}\hyperref[sec:future_directions]{\ref*{sec:future_directions}} & \hspace*{0.8em}\hyperref[sec:future_directions]{\nameref*{sec:future_directions}} & \hyperref[sec:future_directions]{\pageref*{sec:future_directions}} \\
\end{tabularx}\par
\endgroup

\clearpage
\section{Additional Related Work}
\label{sec:extended_related_work}

\subsection{Robot benchmarks and dynamic manipulation}
\label{sec:related_benchmarks}
Existing environments measure transfer, language-conditioned manipulation, and compositional skills~\citep{yu2020meta,james2020rlbench,mees2022calvin,mu2021maniskill}. Other platforms emphasize simulation scale and controlled intervention~\citep{gu2023maniskill2,mendez2022composuite,ahmed2020causalworld}, while household, bimanual, and demonstration resources broaden task and scene diversity~\citep{nasiriany2024robocasa,li2023behavior,chen2025robotwin,walke2023bridgedata}. LIBERO, LIBERO-Plus, and LIBERO-PRO provide controlled studies of transfer, OOD robustness, and memorization~\citep{liu2023libero,fei2025liberoplus,zhou2025liberopro}; LIBERO-Safety evaluates physical and semantic safety~\citep{cui2026liberosafety}. Related stress tests expose visual, embodiment, distractor, and adversarial weaknesses~\citep{pumacay2024colosseum,wang2025vlatest,zhang2024vlabench,li24simpler,zhou2025exploring,liu2025can,fang2025intention,garcia2025towards,kube2025beyond,wang2025exploringadversarialvulnerabilitiesvisionlanguageaction}. These robustness benchmarks normally set external conditions before the first policy action. DynamicVLA and DOMINO directly study moving-object manipulation and low-latency reactive control~\citep{xie2026dynamicvla,fang2026dynamicmanipulation}. Their focus on continuous target motion complements \maxbench, which pairs no-event and event rollouts with identical executed prefixes to test abrupt nuisance, geometry, and path changes.

LIBERO-X~\citep{wang2026liberox} evaluates progressively combined spatial, object, and instruction perturbations through a hierarchy of difficulty levels, and provides diverse human demonstrations collected through teleoperation. This complements \maxbench, which introduces an event after an identical executed action prefix to compare task outcomes with and without a change during execution.

\subsection{Object-centric generalization and changes during execution}
\label{sec:related_objects}
ORION~\citep{zhu_vision_based_manipulation_auro_2026} constructs a manipulation policy from a single human video using object graphs that represent task-relevant states and interaction relations. Its evaluation includes generalization to novel object instances and varied scenes. A complementary extension of \maxbench would test changes to a particular object during an ongoing task while holding its pre-event history fixed.

\subsection{Vision-language-action policies and action commitment}
\label{sec:related_vla}
The VLA lineage ranges from RT-1, RT-2, and PaLM-E~\citep{Brohan2022RT1RT,Brohan2023RT2VM,driess2023palm} to open and generalist policies such as OpenVLA, OpenVLA-OFT, $\pi_0$, $\pi_{0.5}$, FAST, and Octo~\citep{kim2024openvla,kim2025oft,black2025pi0,pi05,Pertsch2025FASTEA,ghosh2024octo}. Recent variants study cognition, diffusion decoding, unified tokenization, geometry, compact policies, and language grounding~\citep{li2024cogact,wen2025diffusionvla,univla,zheng2026xvla,sun2025geovla,zhao2025cotvla,shukor2025smolvla,li2024manipllm}. Complementary work studies whether visual content remains semantically faithful through cross scale reindexing and mechanistic analysis of object hallucination pathways in VLMs~\citep{dong2026generated,liu2026dual}. Methods for progress-aware manipulation and fine-grained spatial VLM reasoning address complementary failure sources by tracking subtask state and improving visual grounding~\citep{liu2026palm,shen2025fine,ye2026datapyramidembodiedmanipulation}. These methods strengthen internal progress or perception, whereas \maxbench tests response to an external change after execution has begun.

\subsection{World models and world-action policies}
\label{sec:related_wam}
World models support visual foresight, model-predictive control, and latent planning~\citep{ha2018world,finn2017_visualforesight,chua2018_pets,hafner2020_dreamer,janner2019_mbpo,schrittwieser2019_muzero,hansen2022_tdmpc,hafner2023_dreamerv3,hansen2024_tdmpc2}. Video foundation models and predictive representations extend these ideas to large-scale spatiotemporal priors~\citep{bruce2024_genie,assran2025vjepa2,agarwal2025cosmos,kong2026driving}. World-action and hybrid policies connect prediction to control through video generation, latent actions, joint training, memory, or causal interleaving~\citep{hu2025video,ma2026dit4dit,Chen_2025_ICCV,cen2025worldvla,bi2025motus,liao2025genie,li2026causal,ye2026gigaworld,sun2026himemwam,li2026lightwam}. Our evaluation measures whether these policies still complete the task after an external event, rather than the accuracy of their predicted video or latent state. The fixed event boundary can also support future adaptive methods that use prediction error to decide when to cancel pending actions and query the policy again.

\section{Benchmark Construction and Intervention Examples}
\label{sec:selection_audit}

This section describes how source tasks become matched Base/Dynamic cases, how events are applied, and which checks determine eligibility. All construction decisions precede policy evaluation.

\subsection{Design requirements}
\label{sec:benchmark_details}

Four requirements define the comparison. \textbf{Temporal placement:} an event occurs after execution begins. \textbf{Pre-event identity:} Base and Dynamic share the task, initial state, instruction, policy seed, and executed action prefix. \textbf{Continued feasibility:} geometry screening checks whether interventions satisfy support, workspace, visibility, and collision constraints. \textbf{Denominator integrity:} cases remain in the score even if they never reach the trigger or receive a later query.

These requirements let us distinguish whether a policy can complete the Base task, reaches the event, and completes the task after the change. The primary score includes every assigned case; paired transitions and coverage diagnostics report what happens at each stage. The camera controls also compare encountering the same changed view at reset and during execution.

\subsection{Source selection and balancing}
\label{sec:case_construction}

A selection cell combines a source category, event type, and parameter variant. Each Plus cell contains 50 unique source tasks, and each PRO cell contains 15 task/initial-state configurations. The resulting 5,600/2,400 split gives every event 700 Plus-derived and 300 PRO-derived cases. Two fixed variants per event probe direction, magnitude, transform, or placement. Table~\ref{tab:selection_contract} summarizes the allocation.

\begin{table*}[!t]
\caption{\textbf{Outcome-independent selection balances all eight events.} Cells combine source category, event, and parameter variant. Every selected case receives one Base and one Dynamic rollout, yielding 16,000 rollouts per policy.}
\label{tab:selection_contract}
\centering

\setlength{\tabcolsep}{5pt}
\footnotesize
\renewcommand{\arraystretch}{1.10}
\begin{tabular*}{\linewidth}{@{\extracolsep{\fill}}lrrrrr@{}}
\toprule
\tableheader Source & \tableheader Categories & \tableheader Events & \tableheader Variants & \tableheader Cases/cell & \tableheader Total cases \\
\midrule
LIBERO-Plus & 7 & 8 & 2 & 50 & 5,600 \\
LIBERO-PRO & 10 & 8 & 2 & 15 & 2,400 \\
\midrule
Combined & 17 & 8 & 2 & --- & 8,000 \\
\bottomrule
\end{tabular*}
\end{table*}

\textbf{Plus-derived cases.} The source catalog contains 10,030 variants across seven categories and five difficulty levels~\citep{fei2025liberoplus}. Image-space events require a valid trigger entity. Relocation requires a movable target or receptacle on planar support. Distractor insertion requires five eligible objects with declared supports, while obstacle insertion requires an unused object sharing the target's support. Capacity-constrained matching fills 112 cells with 50 unique tasks each. The five difficulty levels contribute 1,075, 1,137, 1,135, 1,135, and 1,118 cases, respectively. Event direction, magnitude, object identity, and placement are fixed before evaluation.

\textbf{PRO-derived cases.} We retain semantic, object, position, task, visual noise and glare, camera viewpoint, object texture, view occlusion, object shape, and initial pose categories~\citep{zhou2025liberopro}. Each contains 40 source variants, one for each task in the four LIBERO suites, with ten candidate initial states. Runtime object move is excluded because it already changes the world during execution. The environment category lacks complete scene and initialization assets. Each of the 160 retained cells receives 15 unique task/initial-state configurations, with selection favoring balanced task use. A source task can occur under several events, so configurations are the sampling unit rather than distinct task identities.

We select cases and replace invalid configurations within each cell using only source metadata and simulator checks. Replacements occur before any policy outcomes are observed and preserve all category, event, and parameter quotas. Every policy uses the same final 8,000 cases. The primary micro average weights all cases equally; a secondary 17-category macro average gives equal weight to source categories of different sizes.

\subsection{Events, triggers, and prefix replay}
\label{sec:event_protocol}

The eight events span changes to the target, destination, visual input, clutter, and approach path. Each is applied once and is not reversed by the evaluator before termination. Table~\ref{tab:events} defines the intervention and intended response. Continuing can require perceptual correction even when the task goal remains unchanged.

\begin{table*}[!t]
\caption{\textbf{Eight events test distinct responses to a post-commitment change.} Parameters take one of two fixed variants. \emph{Continue} means that task semantics remain unchanged, although the policy may still need perceptual correction; \emph{replan} means that the target, destination, or path has changed.}
\label{tab:events}
\centering

\setlength{\tabcolsep}{3.5pt}
\footnotesize
\renewcommand{\arraystretch}{1.10}
\renewcommand{\tabularxcolumn}[1]{m{#1}}
\begin{tabularx}{\linewidth}{@{}l l l Y@{}}
\toprule
\tableheader Event & \tableheader Family & \tableheader Expected response & \tableheader Frozen intervention \\
\midrule
Target relocation & Geometry & Replan & Move the task target on its valid support after approach \\
Receptacle relocation & Geometry & Replan & Move the destination while keeping the task feasible \\
Camera shift & Observation & Continue or correct & Shift camera position, yaw, and field of view \\
Illumination switch & Appearance and clutter & Continue & Change scene lighting by a frozen scale \\
Sensor noise onset & Observation & Continue or correct & Add image corruption and a fixed occluded fraction \\
Visual theme switch & Appearance and clutter & Continue or correct & Apply one fixed color and channel transform \\
Distractor burst & Appearance and clutter & Continue & Insert five irrelevant objects on the active support \\
Obstacle insertion & Path constraint & Replan & Insert an object in the target-support approach region \\
\bottomrule
\end{tabularx}
\end{table*}

All events use the deterministic \texttt{pre\_grasp\_proximity} trigger, which fires when the end effector first enters the designated entity's neighborhood during approach. Base records policy queries and actions. Dynamic restores the same initial state, replays the executed prefix, and applies the event at the trigger. The evaluator verifies identical pre-event actions, one event per triggered case, and the intended change after onset.

For chunked policies, the event can arrive while actions predicted before onset remain queued. The number of these queued actions executed after the event is the \emph{stale-action exposure}. Trigger coverage reports the fraction of assigned cases that reach the event; response coverage reports the fraction that receive a subsequent policy query. Both use all assigned cases, as does the primary success rate.

\subsection{Validity and treatment of failures}
\label{sec:validity_checks}

All 5,600 Plus-derived and 2,400 PRO-derived cases pass automated simulator preflight. Checks require a finite, stable state, valid support and workspace placement, visibility, a nonzero observation change, and no new unrelated contact. Scene-construction or rendering failures are repaired before final scoring. All fourteen primary policy evaluations contain outcomes for the same 8,000 cases.

We score every case by its observed task outcome, including cases that never reach the trigger or receive a later query. Coverage reports how often the event occurs and the policy receives a new observation. Automated validity checks do not establish human-perceived naturalness or demonstrate successful completion after every intervention. The human feasibility audit remains incomplete.

\subsection{Qualitative intervention examples}
\label{sec:qualitative_examples}

\begin{figure*}[!t]
\centering
\begin{tabular}{@{}cccc@{}}
\includegraphics[width=0.225\linewidth]{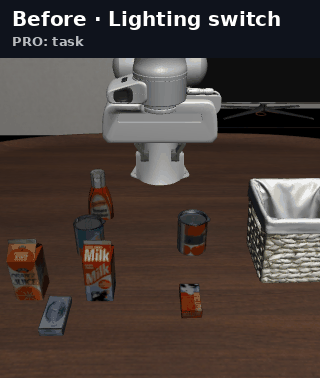} &
\includegraphics[width=0.225\linewidth]{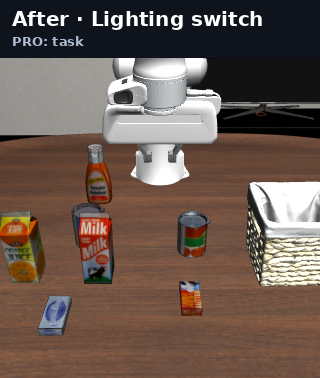} &
\includegraphics[width=0.225\linewidth]{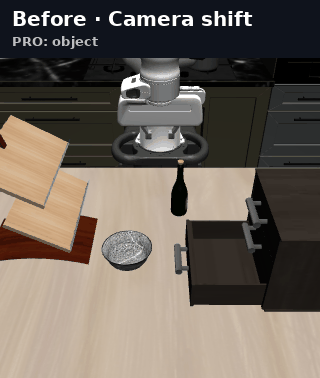} &
\includegraphics[width=0.225\linewidth]{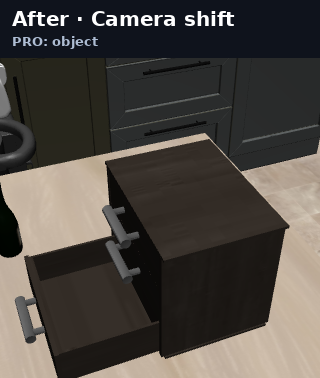} \\
\multicolumn{2}{c}{\textbf{Illumination switch}: Base and Dynamic} & \multicolumn{2}{c}{\textbf{Camera shift}: Base and Dynamic} \\[4pt]
\includegraphics[width=0.225\linewidth]{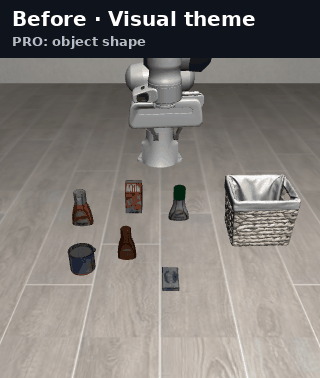} &
\includegraphics[width=0.225\linewidth]{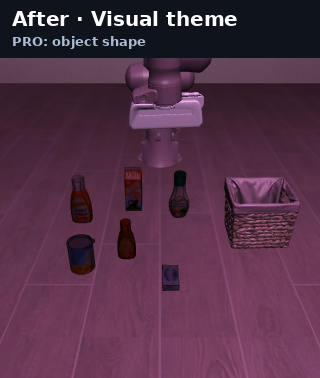} &
\includegraphics[width=0.225\linewidth]{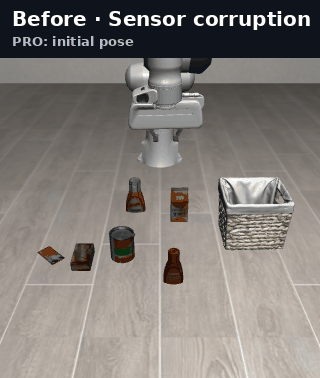} &
\includegraphics[width=0.225\linewidth]{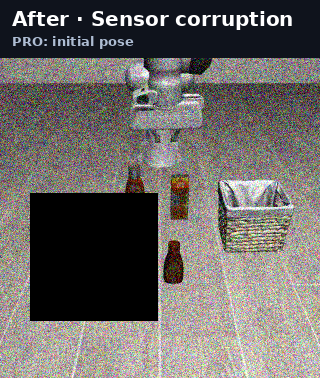} \\
\multicolumn{2}{c}{\textbf{Visual theme switch}: Base and Dynamic} & \multicolumn{2}{c}{\textbf{Sensor corruption}: Base and Dynamic}
\end{tabular}
\caption{\textbf{Paired observation changes.} Each Base image is the exact prechange reference for the adjacent Dynamic image. The intervention changes only the named factor at the frozen trigger.}
\label{fig:qual_obs}
\end{figure*}

\begin{figure*}[!t]
\centering
\begin{tabular}{@{}cccc@{}}
\includegraphics[width=0.225\linewidth]{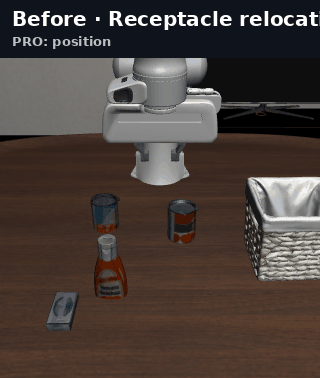} &
\includegraphics[width=0.225\linewidth]{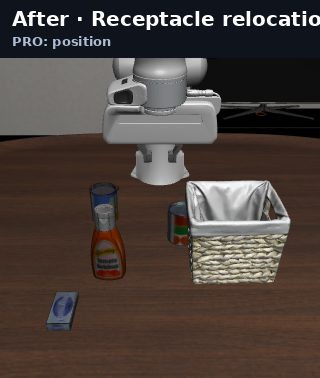} &
\includegraphics[width=0.225\linewidth]{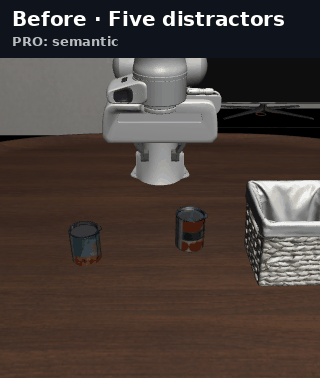} &
\includegraphics[width=0.225\linewidth]{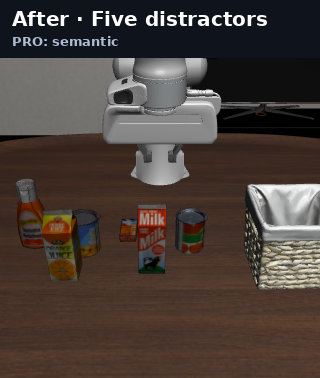} \\
\multicolumn{2}{c}{\textbf{Receptacle relocation}: Base and Dynamic} & \multicolumn{2}{c}{\textbf{Distractor burst}: Base and Dynamic} \\[4pt]
\includegraphics[width=0.225\linewidth]{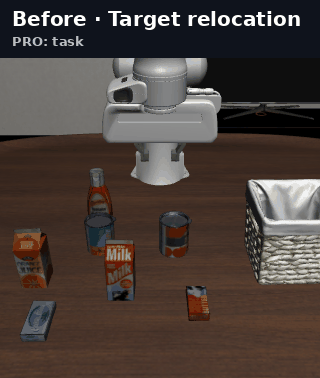} &
\includegraphics[width=0.225\linewidth]{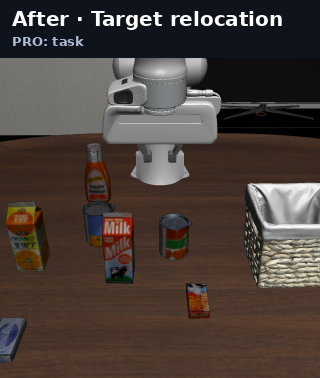} &
\includegraphics[width=0.225\linewidth]{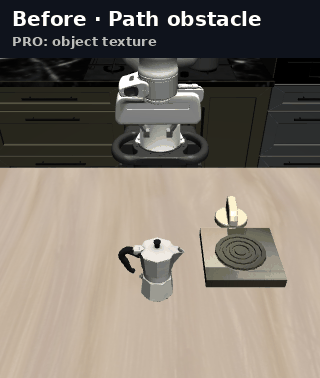} &
\includegraphics[width=0.225\linewidth]{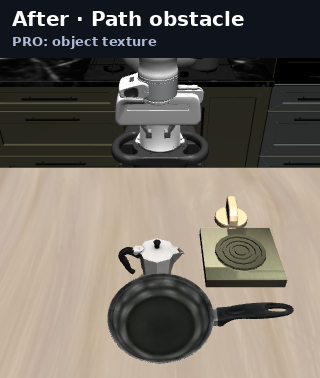} \\
\multicolumn{2}{c}{\textbf{Target relocation}: Base and Dynamic} & \multicolumn{2}{c}{\textbf{Obstacle insertion}: Base and Dynamic}
\end{tabular}
\caption{\textbf{Paired geometry, clutter, and path changes.} Each Base image is the exact prechange reference for the adjacent Dynamic image. The task instruction remains fixed while one declared physical factor changes.}
\label{fig:qual_other}
\end{figure*}

Figs.~\ref{fig:qual_obs} and~\ref{fig:qual_other} show matched Base and Dynamic images for all eight events. Observation events alter visual evidence without moving task objects. Geometry, clutter, and path events modify the physical state and are designed to preserve a feasible completion route.

\section{LIBERO-MAX Lite}
\label{sec:max_lite}

\subsection{Outcome-independent subset}
\label{sec:lite_selection}

LIBERO-MAX Lite uses a fixed subset of 800 released Max case IDs. A persistent pseudorandom generator with seed \texttt{20260830} selects 70 Plus derived and 30 PRO derived cases for each event using only the frozen manifest, event identity, and source label. Selection is independent of checkpoint identity, Base success, Dynamic success, trigger coverage, and query cadence. The resulting Lite track contains 800 matched pairs and 1,600 scored rollouts per checkpoint. This strict subset preserves Max's equal allocation across all eight events and 70:30 source composition. The action cadence analysis uses the same fixed pool.

\subsection{Agreement with the full benchmark}
\label{sec:lite_validation}

Table~\ref{tab:max_lite_validation} compares Lite with Max for fourteen checkpoints whose per-case outcomes can be scored on the fixed subset. All Lite rates use the same 800-pair denominator, so a missing terminal result remains a failure under the benchmark scoring rule. The largest absolute deviation is 2.4 points across all reported Base rates, Dynamic rates, and paired gaps. Lite preserves 88 of 91 pairwise Dynamic orderings. The three differences involve VLA-JEPA versus Xiaomi-Robotics-0 and DiT4DiT versus X-VLA and Light-WAM; these pairs are separated by at most 2.3 points on Max, and DiT4DiT and Light-WAM tie on Lite.

\begin{table*}[!t]
\caption{\textbf{LIBERO-MAX Lite provides a rapid 800-pair estimate of Max on fourteen policies.} Every value is success rate or Dynamic minus Base success in percentage points.}
\label{tab:max_lite_validation}
\centering

\setlength{\tabcolsep}{5pt}
\footnotesize
\renewcommand{\arraystretch}{1.10}
\begin{tabular}{@{}l rr rr rr@{}}
\toprule
\tableheader Model & \tableheader \shortstack{Max\\Base} & \tableheader \shortstack{Lite\\Base} & \tableheader \shortstack{Max\\Dynamic} & \tableheader \shortstack{Lite\\Dynamic} & \tableheader \shortstack{Max\\Gap} & \tableheader \shortstack{Lite\\Gap} \\
\midrule
$\pi_{0.5}$ & 79.7 & 79.3 & 65.7 & 65.1 & $-13.9$ & $-14.1$ \\
OpenVLA-OFT & 64.3 & 64.3 & 43.2 & 45.0 & $-21.1$ & $-19.3$ \\
X-VLA & 62.6 & 62.0 & 37.7 & 39.4 & $-24.9$ & $-22.6$ \\
Xiaomi-Robotics-0 & 70.3 & 70.6 & 52.0 & 53.5 & $-18.3$ & $-17.1$ \\
MolmoAct2 & 80.3 & 79.5 & 66.9 & 68.5 & $-13.4$ & $-11.0$ \\
SmolVLA & 26.1 & 25.6 & 15.0 & 15.3 & $-11.0$ & $-10.4$ \\
GR00T N1.7 & 69.3 & 69.3 & 50.0 & 51.4 & $-19.3$ & $-17.9$ \\
DM0.5 & 79.8 & 79.4 & 62.1 & 63.6 & $-17.7$ & $-15.8$ \\
VLA-JEPA & 73.5 & 72.5 & 54.3 & 53.4 & $-19.2$ & $-19.1$ \\
Cosmos-Policy & 77.4 & 76.1 & 59.3 & 58.6 & $-18.2$ & $-17.5$ \\
Fast-WAM & 42.0 & 41.4 & 24.0 & 24.0 & $-18.0$ & $-17.4$ \\
HiMem-WAM & 73.0 & 72.5 & 57.7 & 58.0 & $-15.3$ & $-14.5$ \\
Light-WAM & 54.8 & 55.3 & 37.3 & 38.5 & $-17.5$ & $-16.8$ \\
DiT4DiT & 65.1 & 65.5 & 39.4 & 38.5 & $-25.7$ & $-27.0$ \\
\bottomrule
\end{tabular}
\end{table*}

\begin{figure*}[!t]
    \centering
    \includegraphics[width=\linewidth]{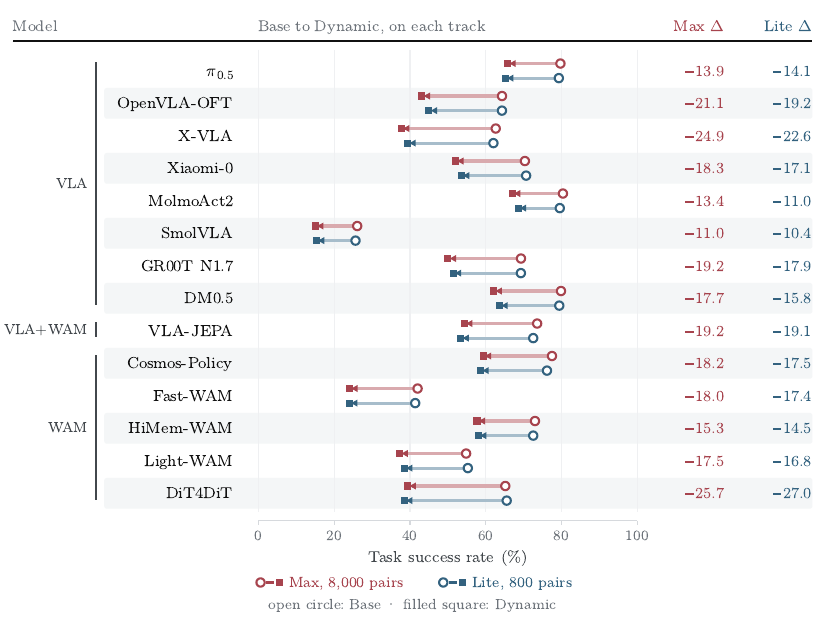}
    \caption{\textbf{The 800-pair Lite track closely reproduces Max across fourteen policies.} Each policy has an upper Max track over 8,000 pairs and a lower Lite track over 800 pairs. Open circles are Base, filled squares Dynamic, and arrows connect them. The two paired changes are printed at right.}
    \label{fig:max_lite_validation}
\end{figure*}

Lite supports integration, debugging, ablation, and early comparison. Max remains the primary reporting track because its larger event cells support more detailed analysis with lower variance. DiT4DiT and DM0.5 each have recorded outcomes for all 800 fixed Lite case IDs; none are reconstructed from aggregate results.

\section{Evaluation Protocol and Policy Details}
\label{sec:model_details}

\subsection{Metrics and statistical comparisons}
\label{sec:protocol_details}

We evaluate all fourteen primary policies on the same 8,000 Base/Dynamic pairs. We report success rates over all assigned cases, paired change $\Delta=\mathrm{SR}_{\mathrm{Dynamic}}-\mathrm{SR}_{\mathrm{Base}}$ in percentage points, and event-family breakdowns. Conditional regression is the fraction of Base-successful cases that fail in Dynamic. We also report retained performance, $\mathrm{SR}_{\mathrm{Dynamic}}/\mathrm{SR}_{\mathrm{Base}}$, and trigger and response coverage. Cases that do not reach the trigger or a post-event query remain in the primary denominator.

Binomial proportions use 95\% intervals, and paired changes use paired bootstrap intervals. Within each policy, the exact McNemar test compares event-associated regression $(1,0)$ with change-associated success $(0,1)$ under the null that these discordant outcomes are equally likely. Cross-policy Dynamic contrasts use common case IDs, with Holm correction across the resulting multiple comparisons. These comparisons describe released policies, whose scale, training data, and serving protocols differ.

We also break down results by event type, task suite, source benchmark, source category, severity, and parameter variant. These comparisons are descriptive unless matched on case ID. Plus-derived and PRO-derived cases are disjoint task instances, so their aggregate difference does not estimate the causal effect of adding PRO. The 17-category macro average gives categories equal weight and remains secondary to the result over all assigned cases. Uncertainty intervals quantify evaluation precision; they do not account for differences in checkpoint training.

\subsection{Released checkpoints and training}
\label{sec:policy_checkpoints}

The primary comparison includes eight VLAs, the predictive VLA-JEPA policy, and five WAMs. Table~\ref{tab:model_details} lists each evaluated LIBERO checkpoint, its training signal, and its official source through the linked model name. Reported scales refer to the principal trainable model or the components named by the model authors; ``not reported'' means that a single comparable total is unavailable.

\begin{table*}[!t]
\caption{\textbf{Released checkpoints cover three policy families.} Scale distinguishes reported totals from separately reported components. The final column summarizes architecture and training of the evaluated LIBERO checkpoint.}
\label{tab:model_details}
\label{tab:models}
\centering

\setlength{\tabcolsep}{3pt}
\scriptsize
\renewcommand{\arraystretch}{1.10}
\renewcommand{\tabularxcolumn}[1]{m{#1}}
\begin{tabularx}{\linewidth}{@{}l >{\raggedright\arraybackslash}m{0.15\linewidth} Y@{}}
\toprule
\tableheader Model & \leavevmode\tableheader Reported scale & \leavevmode\tableheader Backbone and checkpoint training \\
\midrule
\multicolumn{3}{@{}l}{\textit{Vision-language-action policies (VLA)}} \\
\href{https://github.com/Physical-Intelligence/openpi}{$\pi_{0.5}$} & 2B VLM + 300M action expert & PaliGemma with a flow-matching action expert; heterogeneous robot, semantic, and web pretraining followed by LIBERO adaptation~\citep{pi05}. \\
\href{https://github.com/moojink/openvla-oft}{OpenVLA-OFT} & 7B VLA + lightweight heads & Prismatic OpenVLA with continuous regression, parallel decoding, proprioception, and wrist images; combined checkpoint LoRA-optimized on the four standard LIBERO suites~\citep{kim2024openvla,kim2025oft}. \\
\href{https://github.com/2toinf/X-VLA}{X-VLA} & 0.9B & Cross-embodiment VLA with embodiment-specific soft prompts and a flow-matching decoder; released LIBERO checkpoint without MAX-specific adaptation~\citep{zheng2026xvla}. \\
\href{https://github.com/XiaomiRobotics/Xiaomi-Robotics-0}{Xiaomi-Robotics-0} & 5B & Cross-embodiment and vision-language pretraining with asynchronous action-chunk deployment; released LIBERO checkpoint with its native query cadence~\citep{cai2026xiaomirobotics0}. \\
\href{https://github.com/allenai/MolmoAct}{MolmoAct2} & Not reported & Embodied-reasoning VLM with a flow-matching action expert conditioned through per-layer key-value caches; released LIBERO policy with continuous action decoding~\citep{fang2026molmoact2}. \\
\href{https://huggingface.co/HuggingFaceVLA/smolvla_libero}{SmolVLA} & 0.45B variant & SmolVLM2 with flow-matching action generation and asynchronous robot-data training; released \texttt{HuggingFaceVLA/smolvla\_libero} without MAX-specific adaptation~\citep{shukor2025smolvla}. \\
\href{https://huggingface.co/nvidia/GR00T-N1.7-LIBERO}{GR00T N1.7} & 3B & Cosmos-Reason2-2B with proprioception and a flow-matching action transformer; NVIDIA's official LIBERO post-trained checkpoint without MAX-specific adaptation~\citep{bjorck2025gr00t}. \\
\href{https://github.com/dexmal/opendm}{DM0.5} & Not reported & Dexmal's multimodal flow-matching VLA for open-world control; official \texttt{DM05-libero} checkpoint without MAX-specific adaptation. \\
\midrule
\multicolumn{3}{@{}l}{\textit{Predictive VLA (VLA+WAM)}} \\
\href{https://github.com/ginwind/VLA-JEPA}{VLA-JEPA} & 2B VLM + predictive encoders & Qwen3-VL-2B with a V-JEPA2 teacher and latent future-state alignment; human video and single-embodiment robot pretraining followed by end-to-end LIBERO adaptation~\citep{sun2026vlajepa,assran2025vjepa2}. \\
\midrule
\multicolumn{3}{@{}l}{\textit{World-action policies (WAM)}} \\
\href{https://research.nvidia.com/labs/dir/cosmos-policy/}{Cosmos-Policy} & 2B & Cosmos-Predict2-2B adapted without a separate policy backbone; single-stage post-training on target LIBERO demonstrations~\citep{agarwal2025cosmos,kim2026cosmos}. \\
\href{https://github.com/yuantianyuan01/FastWAM}{Fast-WAM} & Approximately 6B & Wan2.2-TI2V-5B plus ActionDiT, jointly trained on video and actions from clean LIBERO demonstrations; inference omits future video generation~\citep{yuan2026fastwam}. \\
\href{https://github.com/Agentic-Intelligence-Lab/HiMem-WAM}{HiMem-WAM} & Not reported & Low-level motion and high-level skill latents with boundary-triggered task memory; released LIBERO checkpoint with native memory-gated inference~\citep{sun2026himemwam}. \\
\href{https://github.com/L1ziang/Light-WAM}{Light-WAM} & 0.44B trainable & Compact video backbone with latent future-video supervision and state-fusion action decoding; released LIBERO checkpoint with its native query cadence~\citep{li2026lightwam}. \\
\href{https://github.com/Mondo-Robotics/DiT4DiT}{DiT4DiT} & 2B video backbone + action DiT & Coupled video-dynamics and action diffusion transformers with dual flow-matching objectives; released LIBERO checkpoint without MAX-specific adaptation~\citep{ma2026dit4dit}. \\
\bottomrule
\end{tabularx}
\end{table*}

\subsection{Action prediction and observation cadence}
\label{sec:policy_execution}

Throughout each primary evaluation, the policy receives new observations and predicts action chunks until the episode ends. The evaluator executes $Q$ actions from each chunk before the next observation and query. Each query returns at most $H$ actions, with $Q\leq H$; when $Q<H$, the policy receives a new observation before all predicted actions have been executed. Table~\ref{tab:execution_protocols} gives the settings used for the primary 8,000-pair comparison.

\begin{table}[!t]
\caption{\textbf{Primary evaluations retain each policy's serving protocol.} $H$ is the maximum number of actions returned by one query; $Q$ is the number executed before the next observation and query.}
\label{tab:execution_protocols}
\centering

\setlength{\tabcolsep}{9pt}
\footnotesize
\renewcommand{\arraystretch}{1.10}
\begin{tabular}{@{}llrr@{}}
\toprule
\tableheader Family & \tableheader Model & \tableheader $H$ & \tableheader $Q$ \\
\midrule
VLA & $\pi_{0.5}$ & 50 & 5 \\
VLA & OpenVLA-OFT & 8 & 8 \\
VLA & X-VLA & 30 & 30 \\
VLA & Xiaomi-Robotics-0 & 30 & 10 \\
VLA & MolmoAct2 & 10 & 10 \\
VLA & SmolVLA & 50 & 10 \\
VLA & GR00T N1.7 & 40 & 8 \\
VLA & DM0.5 & 10 & 10 \\
\midrule
VLA+WAM & VLA-JEPA & 7 & 7 \\
\midrule
WAM & Cosmos-Policy & 16 & 16 \\
WAM & Fast-WAM & 32 & 16 \\
WAM & HiMem-WAM & 32 & 10 \\
WAM & Light-WAM & 32 & 10 \\
WAM & DiT4DiT & 8 & 8 \\
\bottomrule
\end{tabular}
\end{table}

SmolVLA uses ten flow-matching integration steps, and DM0.5 uses ten diffusion integration steps. A separate 200-episode standard-LIBERO integration check for SmolVLA yields 85.0\% success, compared with the 87.3\% reported for its 0.45B variant. The primary $\pi_{0.5}$ setting is $H=50$, $Q=5$; the action-cadence sweep uses a different action-horizon setting, documented with that experiment.

\subsection{Partial-coverage Mimic-Video diagnostic}
\label{sec:mimic_partial}

\href{https://github.com/mimic-video/mimic-video}{Mimic-Video} uses frozen Cosmos-Predict2 video features and a flow-matching inverse-dynamics decoder~\citep{pai2025mimicvideo}. Its public release provides Spatial, Object, and Goal checkpoints but no LIBERO-10 checkpoint. We evaluate all 5,972 supported pairs with the released $H=16$, $Q=5$ protocol. The unsupported 2,028 LIBERO-10 pairs are unavailable, so Mimic-Video is excluded from primary tables, figures, rankings, and cross-model statistics.

\begin{table}[!t]
\caption{\textbf{Mimic-Video loses success across all three supported suites.} Success rates are percentages, and the gap is Dynamic minus Base in percentage points. The supported total contains 5,972 pairs; the 2,028 LIBERO-10 pairs are unavailable. This diagnostic is separate from the primary 8,000-pair comparison.}
\label{tab:mimic_partial}
\centering

\setlength{\tabcolsep}{5pt}
\footnotesize
\renewcommand{\arraystretch}{1.10}
\begin{tabular}{@{}lrrrr@{}}
\toprule
\tableheader Suite & \tableheader Pairs & \tableheader Base SR & \tableheader Dynamic SR & \tableheader Gap \\
\midrule
LIBERO-Goal & 1,656 & 54.2 & 31.4 & $-22.8$ \\
LIBERO-Object & 2,448 & 54.5 & 34.2 & $-20.3$ \\
LIBERO-Spatial & 1,868 & 61.8 & 29.2 & $-32.5$ \\
\midrule
\textbf{Supported total} & \textbf{5,972} & \textbf{56.7} & \textbf{31.9} & \textbf{$-24.8$} \\
LIBERO-10 & 2,028 & N/A & N/A & N/A \\
\bottomrule
\end{tabular}
\end{table}

\section{Paired Outcomes and Robustness Breakdowns}
\label{sec:detailed_results}

We first distinguish cases that fail after the change from those that already fail in Base, then compare events, sources, task suites, and parameter variants. Unless stated otherwise, these results use all fourteen primary policies and the same 8,000 pairs.

\subsection{Paired outcomes and mechanism diagnostics}
\label{sec:paired_diagnostics}

Table~\ref{tab:paired_outcomes_app} separates cases that fail after the event from those that already fail without it. It assigns each pair to one of four outcomes: preserved success, change-associated success, event-associated regression, or persistent failure.

\begin{table*}[!t]
\caption{\textbf{Paired outcomes separate post-change regression from Base competence.} Counts partition the same 8,000 pairs for each of the fourteen primary policies into preserved success, change-associated success, event-associated regression, and persistent failure.}
\label{tab:paired_outcomes_app}
\centering

\footnotesize
\renewcommand{\arraystretch}{1.10}
\begin{tabular*}{\linewidth}{@{\extracolsep{\fill}}lrrrrr@{}}
\toprule
\tableheader Model & \tableheader Preserved & \tableheader Gained & \tableheader Regressed & \tableheader Failed & \tableheader Total \\
 & $(1,1)$ & $(0,1)$ & $(1,0)$ & $(0,0)$ & \\
\midrule
Cosmos-Policy & 4,594 & 148 & 1,601 & 1,657 & 8,000 \\
$\pi_{0.5}$ & 4,998 & 261 & 1,376 & 1,365 & 8,000 \\
OpenVLA-OFT & 3,232 & 220 & 1,909 & 2,639 & 8,000 \\
X-VLA & 2,838 & 177 & 2,171 & 2,814 & 8,000 \\
Xiaomi-Robotics-0 & 3,921 & 237 & 1,703 & 2,139 & 8,000 \\
MolmoAct2 & 5,091 & 264 & 1,336 & 1,309 & 8,000 \\
SmolVLA & 916 & 287 & 1,170 & 5,627 & 8,000 \\
GR00T N1.7 & 3,673 & 328 & 1,868 & 2,131 & 8,000 \\
DM0.5 & 4,757 & 210 & 1,629 & 1,404 & 8,000 \\
VLA-JEPA & 4,121 & 224 & 1,762 & 1,893 & 8,000 \\
Fast-WAM & 1,688 & 234 & 1,672 & 4,406 & 8,000 \\
HiMem-WAM & 4,284 & 329 & 1,553 & 1,834 & 8,000 \\
Light-WAM & 2,723 & 259 & 1,661 & 3,357 & 8,000 \\
DiT4DiT & 3,006 & 147 & 2,205 & 2,642 & 8,000 \\
\bottomrule
\end{tabular*}
\end{table*}

\begin{figure*}[!t]
    \centering
    \includegraphics[width=\linewidth]{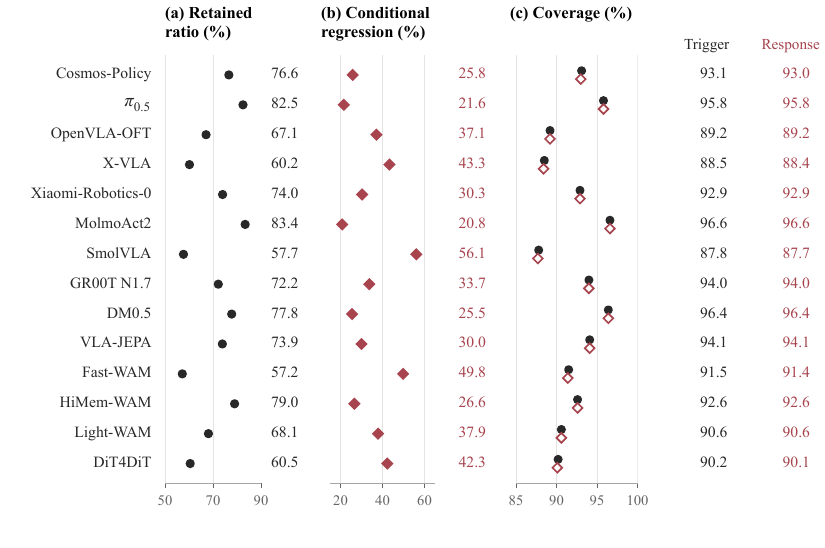}
    \caption{\textbf{Most cases reach the event and a later policy query, but many Base successes are lost.} For each of fourteen policies, retained ratio is Dynamic success rate divided by Base success rate, including change-associated successes; conditional regression is the fraction of Base successes that become Dynamic failures. Trigger and response coverage each use all 8,000 evaluated cases per policy; response requires a policy query after the event. Upper filled circles show trigger coverage and lower hollow diamonds show response coverage. All values are percentages, and the three panels use separate horizontal scales.}
    \label{fig:mechanism_diagnostics}
\end{figure*}

MolmoAct2 retains the most successes, DiT4DiT has the most event-associated regressions, and SmolVLA has the most persistent failures. Fig.~\ref{fig:mechanism_diagnostics} reports these results relative to Base performance, alongside trigger and response coverage. Coverage ranges from 87.7\% for SmolVLA to 96.6\% for MolmoAct2. Cases that fail before the event or before a later query still contribute to the primary success rate.

The online event breaks 20.8\% to 56.1\% of Base successes across the fourteen checkpoints. SmolVLA has the smallest overall gap, at 11.0 points, but loses the largest fraction of Base successes, at 56.1\%; only 26.1\% of its Base episodes succeed. MolmoAct2 at 20.8\%, $\pi_{0.5}$ at 21.6\%, DM0.5 at 25.5\%, Cosmos-Policy at 25.8\%, and HiMem-WAM at 26.6\% have the lowest conditional regression rates.

\subsection{Event-level robustness}
\label{sec:event_diagnostics}

Policies lose less success under some nuisance changes than under changes to perception, the task target, or the path (Fig.~\ref{fig:events}; full values in Table~\ref{tab:event_deltas_app}). Illumination and visual-theme changes cause comparatively small losses, consistent with stability under nuisance appearance changes. Target or receptacle relocation causes the largest loss for eleven policies. Fast-WAM instead falls from 43.4\% Base success to 0.3\% under sensor-noise onset, a 43.1-point drop, while X-VLA is most affected by camera shift. Camera and geometry changes remain consistently harmful, even though they call for different responses: perceptual correction and physical replanning. Final outcomes show how sensitivity varies by event; they do not reveal whether a policy detected the event internally or attempted the intended response.

Policies tend to struggle with the same events. Across the eight event losses, the mean pairwise Spearman correlation is $0.80$ over the fourteen policies, although individual pairs range from $0.29$ to $1.00$. Geometry losses range from 21.2 to 52.4 points and observation losses range from 13.3 to 49.7 points. Both are generally larger than appearance-and-clutter losses of 1.2 to 16.2 points and path-constraint losses of 1.8 to 8.3 points. Sensor noise causes the largest losses for Fast-WAM and DiT4DiT, camera shift for X-VLA; HiMem-WAM remains unusually robust to camera changes. GR00T N1.7 and DM0.5 are particularly sensitive to geometry changes: target relocation reduces their success by 52.5 and 39.9 points, respectively, while appearance-and-clutter losses remain 2.2 and 5.2 points. These differences show why event-specific results are more informative than a family-wide ranking.

Table~\ref{tab:event_deltas_app} reports paired losses for all eight events. Each event receives equal weight, so comparing rows requires no correction for event frequency.

\begin{table*}[!t]
\caption{\textbf{Event sensitivity varies across policies.} Entries show Dynamic minus Base success in percentage points for every policy and event. Every row has 1,000 pairs per event.}
\label{tab:event_deltas_app}
\centering

\setlength{\tabcolsep}{3.5pt}
\scriptsize
\renewcommand{\arraystretch}{1.10}
\renewcommand{\tabularxcolumn}[1]{m{#1}}
\begin{tabularx}{\linewidth}{@{}Y rrrrrrrr@{}}
\toprule
\tableheader Model & \tableheader Camera & \tableheader Distractor & \tableheader Light & \tableheader Obstacle & \tableheader Receptacle & \tableheader Sensor & \tableheader Target & \tableheader Theme \\
\midrule
Cosmos-Policy & $-23.9$ & $-15.7$ & $-0.3$ & $-4.6$ & $-24.1$ & $-26.4$ & $-47.5$ & $-2.8$ \\
$\pi_{0.5}$ & $-23.9$ & $-10.9$ & $+0.2$ & $-4.2$ & $-13.8$ & $-22.3$ & $-34.6$ & $-2.0$ \\
OpenVLA-OFT & $-35.6$ & $-26.3$ & $-1.2$ & $-6.8$ & $-25.1$ & $-28.3$ & $-42.1$ & $-3.5$ \\
X-VLA & $-55.9$ & $-3.5$ & $+0.7$ & $-5.1$ & $-52.4$ & $-30.0$ & $-52.3$ & $-0.9$ \\
Xiaomi-Robotics-0 & $-30.6$ & $-15.4$ & $-0.1$ & $-3.9$ & $-31.3$ & $-19.8$ & $-44.0$ & $-1.5$ \\
MolmoAct2 & $-17.9$ & $-13.9$ & $+0.6$ & $-1.8$ & $-21.1$ & $-20.6$ & $-32.0$ & $-0.5$ \\
SmolVLA & $-14.7$ & $-3.9$ & $-2.3$ & $-2.7$ & $-18.6$ & $-18.5$ & $-23.9$ & $-3.7$ \\
GR00T N1.7 & $-31.8$ & $-6.1$ & $+0.3$ & $-8.3$ & $-37.5$ & $-17.3$ & $-52.5$ & $-0.8$ \\
DM0.5 & $-33.6$ & $-12.3$ & $-0.9$ & $-5.7$ & $-22.7$ & $-24.4$ & $-39.9$ & $-2.4$ \\
VLA-JEPA & $-31.1$ & $-15.6$ & $-2.9$ & $-4.0$ & $-26.6$ & $-29.4$ & $-40.7$ & $-3.5$ \\
Fast-WAM & $-28.1$ & $-3.7$ & $-3.6$ & $-6.1$ & $-24.5$ & $-43.1$ & $-30.4$ & $-4.3$ \\
HiMem-WAM & $+0.9$ & $-13.1$ & $+0.1$ & $-3.3$ & $-35.8$ & $-27.5$ & $-43.7$ & $0.0$ \\
Light-WAM & $-3.0$ & $-8.6$ & $-0.9$ & $-2.0$ & $-39.1$ & $-40.4$ & $-40.7$ & $-5.5$ \\
DiT4DiT & $-31.5$ & $-16.6$ & $-7.6$ & $-4.5$ & $-18.6$ & $-67.9$ & $-34.7$ & $-24.4$ \\
\bottomrule
\end{tabularx}
\end{table*}

\subsection{Robustness across source categories}
\label{sec:source_diagnostics}

\textbf{Source subsets and low Base success.}
All fourteen evaluated checkpoints have lower Dynamic success and a smaller paired drop on PRO derived cases than on Plus derived cases (Table~\ref{tab:substrate}). The two subsets contain different task instances, and lower PRO derived Base success leaves less room to fall. Their aggregate difference therefore does not establish that PRO increases sensitivity to online events.

\begin{table*}[!t]
\caption{\textbf{Success falls after the change on both Plus and PRO cases.} Plus contains 7 categories and 5,600 cases; PRO contains 10 categories and 2,400 cases. Base and Dynamic success rates use all assigned cases, showing where low Base success limits the size of the loss. The final columns identify the source categories with the lowest Base success and the largest loss.}
\label{tab:substrate}
\centering

\setlength{\tabcolsep}{2pt}
\scriptsize
\renewcommand{\arraystretch}{1.10}
\begin{tabular*}{\linewidth}{@{\extracolsep{\fill}}llrrrll@{}}
\toprule
\tableheader Model & \tableheader Source & \tableheader Base & \tableheader Dynamic & \tableheader Gap & \tableheader Lowest Base & \tableheader Largest loss \\
\midrule
& Plus & 83.0 & 63.2 & $-19.8$ & Robot initial state & Background texture \\
\multirow[c]{-2}{*}{Cosmos-Policy} & PRO & 64.5 & 50.2 & $-14.4$ & Position & Noise and glare \\
& Plus & 86.1 & 70.3 & $-15.8$ & Camera viewpoint & Light condition \\
\multirow[c]{-2}{*}{$\pi_{0.5}$} & PRO & 64.8 & 55.1 & $-9.7$ & Initial pose & Noise and glare \\
& Plus & 68.3 & 45.2 & $-23.0$ & Robot initial state & Light condition \\
\multirow[c]{-2}{*}{OpenVLA-OFT} & PRO & 55.0 & 38.3 & $-16.7$ & Initial pose & Noise and glare \\
& Plus & 67.9 & 41.2 & $-26.6$ & Camera viewpoint & Background texture \\
\multirow[c]{-2}{*}{X-VLA} & PRO & 50.4 & 29.5 & $-20.9$ & Camera angle & Noise and glare \\
& Plus & 74.5 & 54.4 & $-20.1$ & Camera viewpoint & Sensor noise \\
\multirow[c]{-2}{*}{Xiaomi-Robotics-0} & PRO & 60.5 & 46.4 & $-14.1$ & Position & Noise and glare \\
& Plus & 86.6 & 71.8 & $-14.9$ & Robot initial state & Sensor noise \\
\multirow[c]{-2}{*}{MolmoAct2} & PRO & 65.7 & 55.7 & $-10.0$ & Camera angle & Object shape \\
& Plus & 26.8 & 15.4 & $-11.4$ & Sensor noise & Light condition \\
\multirow[c]{-2}{*}{SmolVLA} & PRO & 24.4 & 14.3 & $-10.1$ & Position & Object shape \\
& Plus & 75.3 & 54.1 & $-21.2$ & Robot initial state & Light condition \\
\multirow[c]{-2}{*}{GR00T N1.7} & PRO & 55.2 & 40.5 & $-14.7$ & Initial pose & Semantic \\
& Plus & 86.4 & 66.9 & $-19.5$ & Camera viewpoint & Light condition \\
\multirow[c]{-2}{*}{DM0.5} & PRO & 64.6 & 50.9 & $-13.7$ & Initial pose & Noise and glare \\
& Plus & 79.0 & 58.3 & $-20.8$ & Sensor noise & Background texture \\
\multirow[c]{-2}{*}{VLA-JEPA} & PRO & 60.8 & 45.1 & $-15.6$ & Initial pose & Noise and glare \\
& Plus & 42.8 & 24.4 & $-18.4$ & Camera viewpoint & Light condition \\
\multirow[c]{-2}{*}{Fast-WAM} & PRO & 40.2 & 23.1 & $-17.0$ & Noise and glare & Object shape \\
& Plus & 76.8 & 60.5 & $-16.3$ & Robot initial state & Background texture \\
\multirow[c]{-2}{*}{HiMem-WAM} & PRO & 64.0 & 51.1 & $-12.9$ & Initial pose & Noise and glare \\
& Plus & 56.2 & 38.3 & $-17.9$ & Robot initial state & Light condition \\
\multirow[c]{-2}{*}{Light-WAM} & PRO & 51.5 & 34.8 & $-16.6$ & Position & Object shape \\
& Plus & 70.7 & 42.5 & $-28.2$ & Background texture & Light condition \\
\multirow[c]{-2}{*}{DiT4DiT} & PRO & 52.2 & 32.2 & $-20.0$ & Initial pose & Object texture \\
\bottomrule
\end{tabular*}
\end{table*}

\begin{figure*}[!t]
    \centering
    \includegraphics[width=\linewidth]{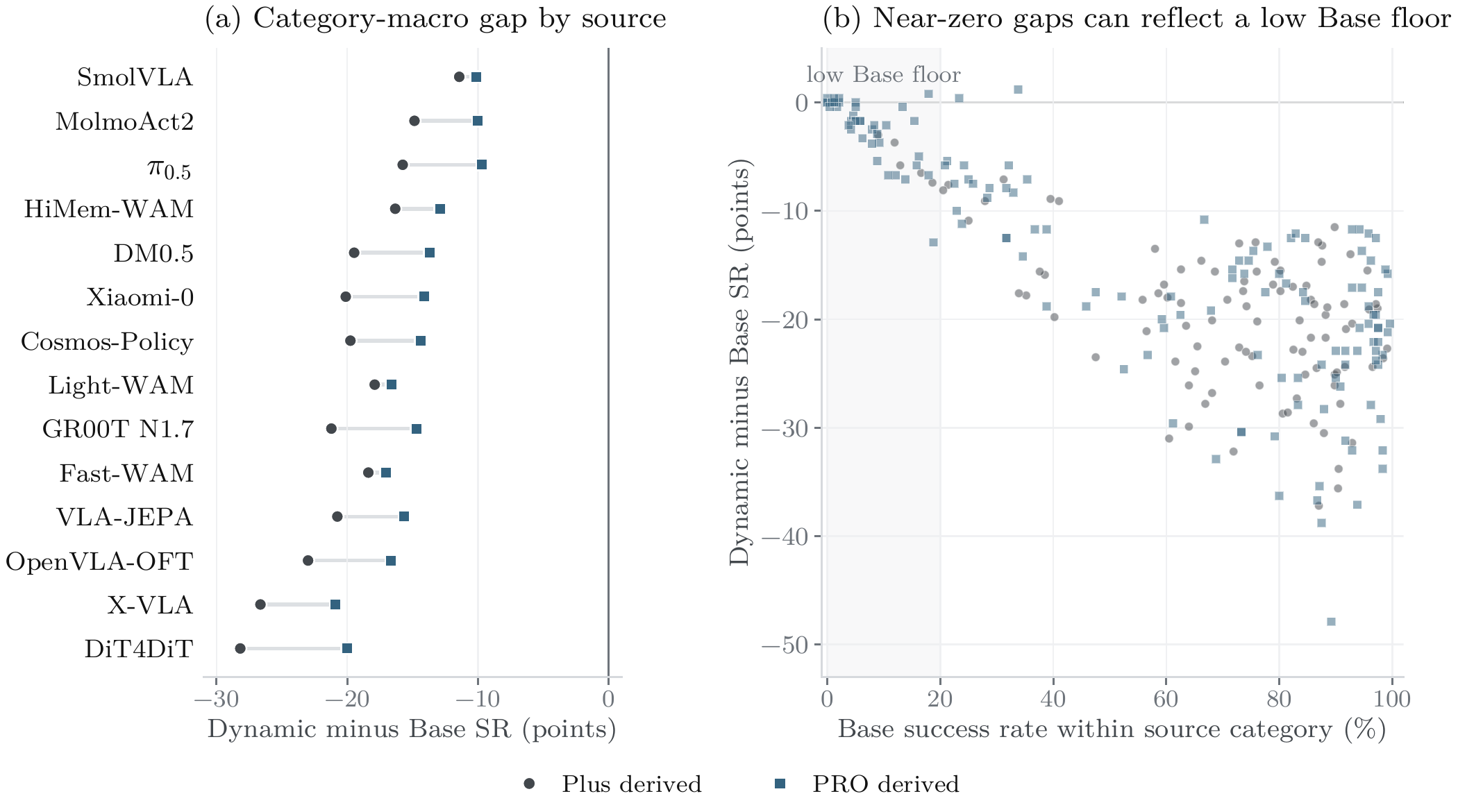}
    \caption{\textbf{Paired change across all 17 source categories.} (a) Category-macro Dynamic minus Base success for the seven Plus categories (circles) and ten PRO categories (squares). (b) Each model-category gap plotted against its Base success: near-zero gaps cluster where Base success is already below 20\%, leaving little room for success to fall.}
    \label{fig:substrate_heatmap}
\end{figure*}

\textbf{Source-category patterns.}
Fig.~\ref{fig:substrate_heatmap} shows paired changes across all 17 source categories. Every policy loses success across many Plus derived categories; several PRO derived categories already have low Base success, which limits how far success can fall. Policies also differ in which categories cause the largest losses: X-VLA loses 37.2 points on Plus background texture and 47.9 on PRO noise and glare, whereas MolmoAct2 has smaller aggregate losses but remains sensitive to sensor noise and object shape.

\begin{figure*}[!t]
\centering
\includegraphics[width=\linewidth]{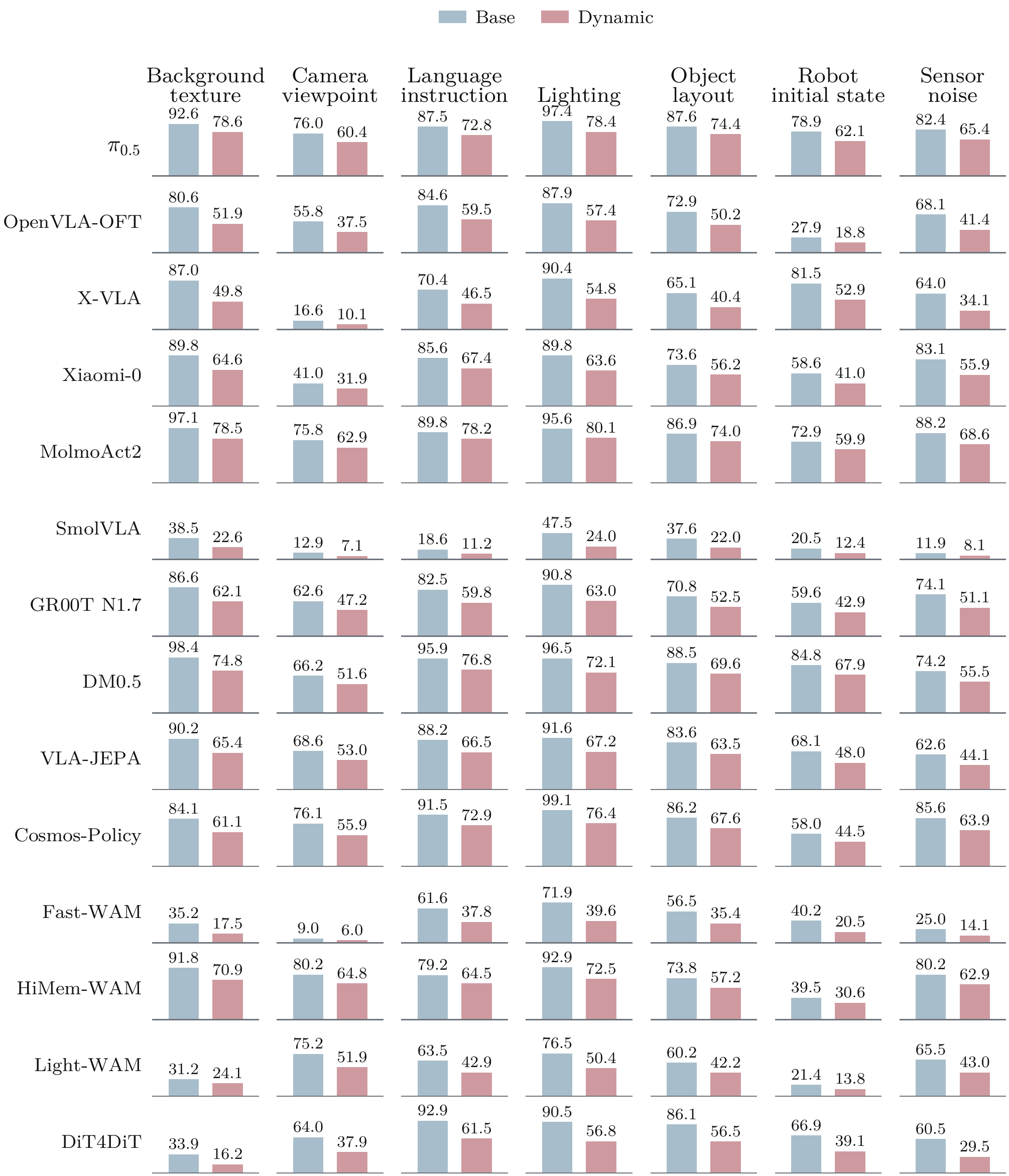}
\caption{\textbf{Base and Dynamic success across the seven Plus derived source categories.} Most of the 98 model-category comparisons show losses; some categories already have low Base success. Every model-category cell contains 800 pairs.}
\label{fig:all_categories_plus}
\end{figure*}

\begin{figure*}[!t]
\centering
\includegraphics[width=\linewidth]{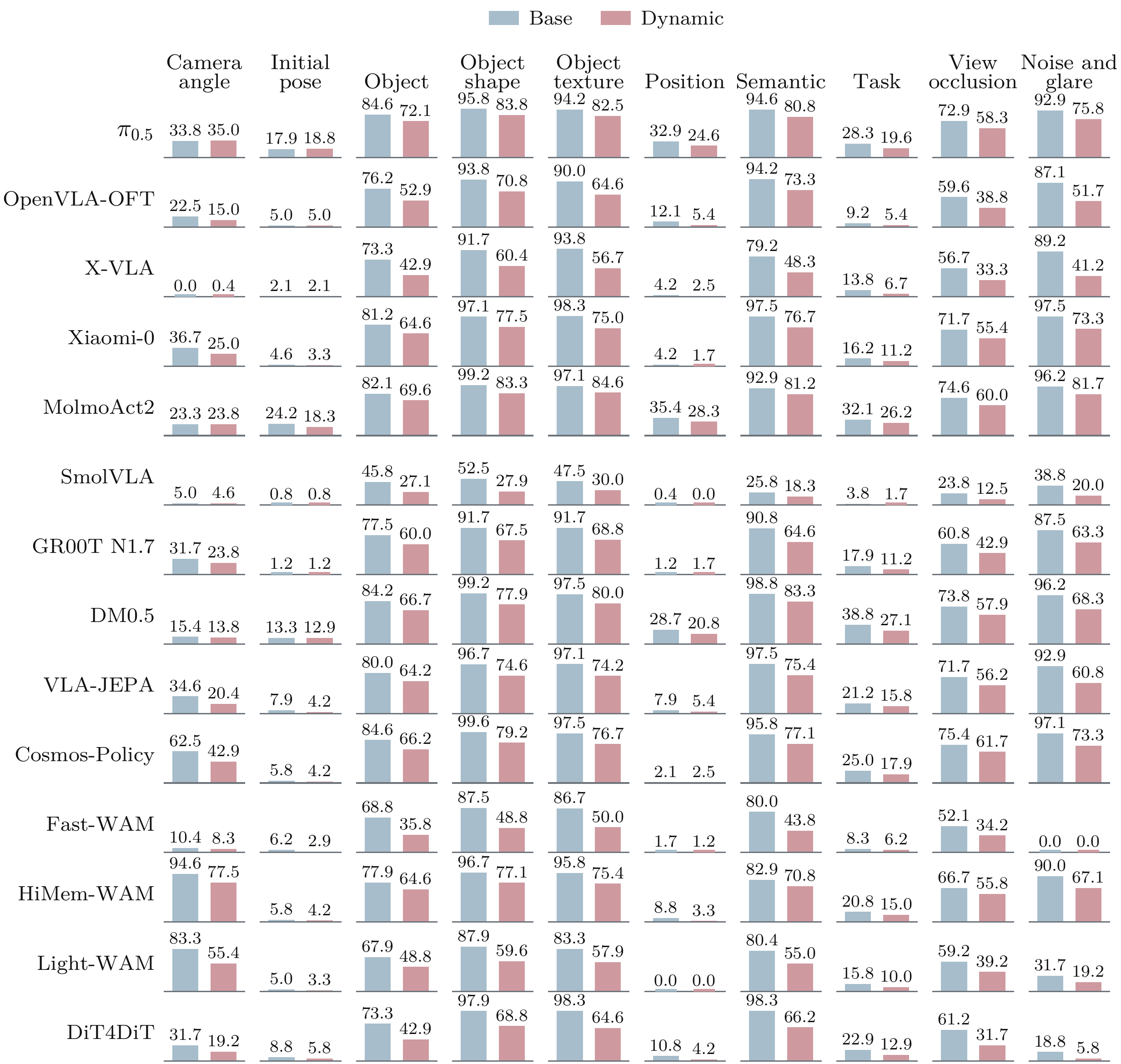}
\caption{\textbf{Base and Dynamic success across the ten PRO derived source categories.} Each category contains 240 matched pairs. Low Base success explains the small or positive differences for several pose, position, and task settings.}
\label{fig:all_categories_pro}
\end{figure*}

A small or positive gap can coexist with frequent failures when Base success is already low. For example, SmolVLA scores only 0.4\% on the PRO position Base cases, and Fast-WAM scores 0\% on visual noise and glare in both trajectories. Figs.~\ref{fig:all_categories_plus} and~\ref{fig:all_categories_pro} therefore show both Base and Dynamic success alongside the heatmap of paired changes. The average loss remains negative when all 17 source categories receive equal weight: macro deltas range from $-10.7$ to $-23.4$ across all fourteen checkpoints.

Figs.~\ref{fig:all_categories_plus} and~\ref{fig:all_categories_pro} compare Base and Dynamic success directly for every model and source category. The paired bars show how much success falls and how often the policy succeeds in Base. A small gap supports a claim of robustness only when the policy can already solve the Base tasks.

Across 98 Plus source comparisons, most gaps are negative, with a small number of checkpoint-specific exceptions. Depending on the checkpoint, the largest Plus loss occurs in background texture, light condition, sensor noise, or language. PRO results vary more across categories. Several have Base success below 20\%, so small positive gaps do not establish a successful post-change response. Categories with high Base success show large losses on object shape, object texture, semantics, and visual noise and glare.

\subsection{Task suites, severity metadata, and parameter sensitivity}
\label{sec:factorized_diagnostics}

Table~\ref{tab:factorized} tests whether the main gap is concentrated in one task suite or one severity label. It also records the largest difference between the two frozen parameter variants for each checkpoint.

\begin{table*}[!t]
\caption{\textbf{Losses persist across task suites, with sensitivity to event parameters.} All gaps are Dynamic minus Base success in percentage points. Category macro weights all 17 source categories equally. Medium and high contain 1,000 and 7,000 cases with different event compositions. Split is the largest within-event Dynamic success difference between two variant cells.}
\label{tab:factorized}
\centering

\setlength{\tabcolsep}{4pt}
\scriptsize
\renewcommand{\arraystretch}{1.10}
\textbf{Task-suite and category gaps}\\[3pt]
\begin{tabular*}{\linewidth}{@{\extracolsep{\fill}}lrrrrr@{}}
\toprule
\tableheader Model & \tableheader Macro & \tableheader LIBERO-10 & \tableheader Goal & \tableheader Object & \tableheader Spatial \\
\midrule
Cosmos-Policy & $-16.6$ & $-22.9$ & $-19.8$ & $-12.9$ & $-18.5$ \\
$\pi_{0.5}$ & $-12.2$ & $-21.2$ & $-14.7$ & $-10.4$ & $-10.1$ \\
OpenVLA-OFT & $-19.3$ & $-22.3$ & $-21.4$ & $-21.8$ & $-18.6$ \\
X-VLA & $-23.3$ & $-30.1$ & $-24.6$ & $-17.2$ & $-29.7$ \\
Xiaomi-Robotics-0 & $-16.6$ & $-23.5$ & $-17.0$ & $-13.8$ & $-19.8$ \\
MolmoAct2 & $-12.0$ & $-16.0$ & $-15.0$ & $-9.6$ & $-14.1$ \\
SmolVLA & $-10.7$ & $-10.9$ & $-5.4$ & $-12.5$ & $-14.1$ \\
GR00T N1.7 & $-17.4$ & $-25.6$ & $-16.8$ & $-14.5$ & $-20.8$ \\
DM0.5 & $-16.1$ & $-25.1$ & $-17.1$ & $-13.2$ & $-16.2$ \\
VLA-JEPA & $-17.7$ & $-24.1$ & $-18.7$ & $-14.1$ & $-21.3$ \\
Fast-WAM & $-17.6$ & $-16.0$ & $-17.0$ & $-16.6$ & $-22.8$ \\
HiMem-WAM & $-14.3$ & $-17.1$ & $-13.5$ & $-11.8$ & $-19.5$ \\
Light-WAM & $-17.2$ & $-12.1$ & $-17.0$ & $-16.4$ & $-25.3$ \\
DiT4DiT & $-23.4$ & $-26.3$ & $-30.7$ & $-18.3$ & $-30.5$ \\
\bottomrule
\end{tabular*}
\par\medskip
\textbf{Severity groups and parameter sensitivity}\\[3pt]
\begin{tabular*}{\linewidth}{@{\extracolsep{\fill}}lrrlr@{}}
\toprule
\tableheader Model & \tableheader Medium & \tableheader High & \tableheader Largest-split event & \tableheader Split (pp) \\
\midrule
Cosmos-Policy & $-20.6$ & $-17.8$ & Receptacle & 31.0 \\
$\pi_{0.5}$ & $-13.5$ & $-14.0$ & Camera & 35.6 \\
OpenVLA-OFT & $-20.0$ & $-21.3$ & Target & 27.6 \\
X-VLA & $-41.1$ & $-22.6$ & Receptacle & 28.4 \\
Xiaomi-Robotics-0 & $-20.2$ & $-18.1$ & Target & 35.6 \\
MolmoAct2 & $-14.4$ & $-13.3$ & Sensor & 31.6 \\
SmolVLA & $-16.7$ & $-10.2$ & Receptacle & 12.2 \\
GR00T N1.7 & $-32.2$ & $-17.4$ & Receptacle & 35.6 \\
DM0.5 & $-16.1$ & $-18.0$ & Target & 34.8 \\
VLA-JEPA & $-18.2$ & $-19.4$ & Target & 34.6 \\
Fast-WAM & $-17.4$ & $-18.1$ & Receptacle & 20.6 \\
HiMem-WAM & $-23.1$ & $-14.2$ & Receptacle & 36.4 \\
Light-WAM & $-31.2$ & $-15.6$ & Receptacle & 23.0 \\
DiT4DiT & $-16.9$ & $-27.0$ & Target & 25.8 \\
\bottomrule
\end{tabular*}
\end{table*}

\textbf{Event parameters and task suites.}
Success can differ substantially between the two fixed parameter variants of the same event. Across fourteen checkpoints, the largest within-event Dynamic SR split ranges from 12.2 to 36.4 points. A benchmark's choice of event parameters can therefore materially affect measured robustness. \maxbench balances both variants and identifies them in the results as parameter variants, not independent model seeds.

The largest losses also occur in different task suites for different policies. LIBERO-10 produces the largest drop for several checkpoints, including GR00T N1.7, while LIBERO-Spatial is largest for Fast-WAM, HiMem-WAM, Light-WAM, and SmolVLA; DiT4DiT is most affected on LIBERO-Goal. The 17-category macro deltas remain negative for all fourteen evaluated checkpoints, so the overall loss persists when each source category receives equal weight. Individual PRO categories with Base success near zero still require care: reporting Base and Dynamic results together makes these floor effects visible.

\section{Camera Controls: Viewpoint and Execution History}
\label{sec:camera_timing_control}

\subsection{Design and paired comparisons}
\label{sec:camera_design}
A new camera view may be difficult even when the policy sees it from the start. We compare that setting with introducing the same view during an ongoing task, asking how viewpoint difficulty and the preceding trajectory affect task completion. X-VLA, $\pi_{0.5}$, and HiMem-WAM share all 1,000 camera cases from the frozen Max benchmark: 700 Plus-derived and 300 PRO-derived cases, selected without policy outcomes. Base, Reset, and Mid-task give 9,000 complete rollouts across the three models. This diagnostic preserves the original 8,000-pair benchmark and its primary results.

\textbf{Three conditions.}
Base retains each case's original source configuration without an additional camera event. Reset restores that source configuration and initial state, then applies the complete camera transform before the first policy input; the changed camera remains active until the episode ends. Mid-task uses the original Dynamic protocol: replay the matched Base action prefix up to the original proximity trigger, apply the same transform, and continue with the native action queue. Reset generates its own actions from the changed observations. Within each model, the conditions share the case, initial state, policy seed 195, source runtime, step budget, and success criterion.

X-VLA retains $H=Q=30$, $\pi_{0.5}$ retains $H=10$, $Q=5$, and HiMem-WAM retains $H=32$, $Q=10$ with ten inference steps. Checkpoints and action interfaces are fixed across conditions. All 3,000 model--case triples pass the initial-state, executed-prefix, and camera checks. Executed Base/Mid prefixes match exactly before the event, and Reset's persistent camera position, orientation, and field of view match Mid-task in every triggered case. Trigger and subsequent-query counts are 916 and 915 for X-VLA, 951 and 951 for $\pi_{0.5}$, and 941 and 941 for HiMem-WAM. Every model retains all 1,000 cases in its primary denominator, including unreached events.

\subsection{Timing effects and source consistency}
\label{sec:camera_effects}
\begin{figure*}[!t]
    \centering
    \includegraphics[width=\linewidth]{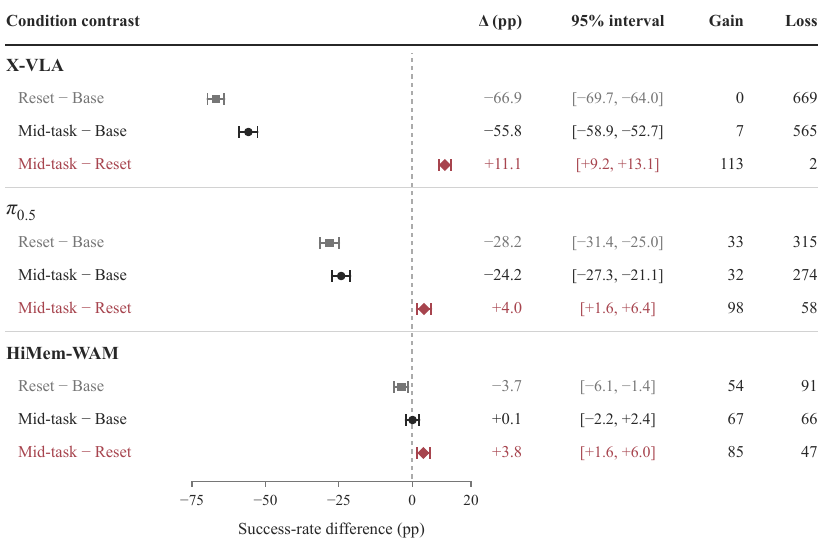}
    \caption{\textbf{Mid-task success exceeds Reset for all three policies.} Points show paired success-rate differences, with pointwise 95\% empirical bootstrap intervals, over the same 1,000 camera cases per model; resampling retains the 700 Plus / 300 PRO composition. Gray squares denote Reset minus Base, black circles Mid-task minus Base, and red diamonds Mid-task minus Reset. Gain and Loss count cases where the first condition succeeds and the second fails, or vice versa.}
    \label{fig:camera_paired_contrasts}
\end{figure*}

The changed view reduces success even when present from the first input (Table~\ref{tab:camera_timing_success}). X-VLA achieves 68.2\% in Base, 1.3\% in Reset, and 12.4\% in Mid-task; $\pi_{0.5}$ achieves 79.5\%, 51.3\%, and 55.3\%, respectively. HiMem-WAM achieves 72.5\%, 68.8\%, and 72.6\%. Mid-task exceeds Reset by 11.1, 4.0, and 3.8 points (paired 95\% intervals in Fig.~\ref{fig:camera_paired_contrasts}). All three policies therefore succeed more often when they first make progress in the original view, although X-VLA and $\pi_{0.5}$ still struggle with the changed view.

For each contrast, we average within-case differences in binary success. Resampling retains the 700/300 source composition and all three paired conditions. We compute the 2.5th and 97.5th percentiles of the empirical bootstrap distribution by discrete convolution.

\textbf{Source and task sensitivity.}
Mid-task success exceeds Reset success on both sources: by $+11.6/+10.0$ points for X-VLA, $+4.1/+3.7$ for $\pi_{0.5}$, and $+4.7/+1.7$ for HiMem-WAM on Plus/PRO, respectively. The PRO-only intervals for $\pi_{0.5}$ and HiMem-WAM include zero. Table~\ref{tab:camera_timing_sources} reports all source counts.

The 1,000 cases cover 995 source/task/initial-configuration groups and 40 reference tasks. We also resample the 40 reference tasks jointly across sources, include all cases belonging to each sampled task, and combine source-wise pooled success rates with fixed 0.7/0.3 weights. Across 50,000 replicates, the paired Mid-task minus Reset intervals are $[6.4,17.0]$, $[0.8,7.2]$, and $[0.8,7.2]$ points for X-VLA, $\pi_{0.5}$, and HiMem-WAM. Resampling the 995 configuration groups or 80 source-specific reference-task groups independently within source supports the same Mid-task advantage. These checks keep the original case-weighted point estimates and test whether the interval estimates depend on how cases are grouped. They cover the observed cohort; additional policy seeds and training runs remain outside this analysis.

\begin{table*}[t]
    \caption{\textbf{Camera-control outcomes vary across source partitions.} Cells are success counts, with the source-specific denominator in the third column. The same cases appear in all three conditions.}
    \label{tab:camera_timing_sources}
    \centering

    \setlength{\tabcolsep}{7pt}
    \footnotesize
\renewcommand{\arraystretch}{1.10}
\begin{tabular}{@{}llrrrr@{}}
    \toprule
\tableheader Model & \tableheader Source & \tableheader Cases & \tableheader Base & \tableheader Reset & \tableheader Mid-task \\
\midrule
    \multirow[c]{2}{*}{X-VLA} & Plus & 700 & 524 & 12 & 93 \\
     & PRO & 300 & 158 & 1 & 31 \\
    \midrule
    \multirow[c]{2}{*}{$\pi_{0.5}$} & Plus & 700 & 606 & 382 & 411 \\
     & PRO & 300 & 189 & 131 & 142 \\
    \midrule
    \multirow[c]{2}{*}{HiMem-WAM} & Plus & 700 & 531 & 506 & 539 \\
     & PRO & 300 & 194 & 182 & 187 \\
    \bottomrule
    \end{tabular}
\end{table*}

\subsection{Paired outcomes and trajectory diagnostics}
\label{sec:camera_trajectories}
HiMem-WAM's near-zero Base-to-Mid difference combines 66 cases that succeed in Base but fail in Mid-task with 67 that fail in Base but succeed in Mid-task. Among cases where both Base and Reset succeed, Mid-task fails in 2/13 X-VLA cases, 48/480 $\pi_{0.5}$ cases, and 34/634 HiMem-WAM cases. Thus, each policy has individual Mid-task failures despite its higher aggregate success than Reset. These groups are selected from each policy's observed successes; they do not independently establish that the policy can reliably handle the new view.

The two comparisons answer different questions. Base versus Mid-task measures the effect of adding a camera event after the same executed history. Reset versus Mid-task compares seeing the changed view from the start with seeing it after making progress in the original view. This second comparison also changes how long the policy sees the new view, which states it visits, and how many steps remain. The results measure whether the policy completes the task under these conditions; identifying internal detection or deliberate replanning would require additional instrumentation.

\textbf{Approach and post-event behavior.}
We use the trigger's 18\,cm Euclidean end-effector distance to the designated entity as an approach diagnostic. Reconstructing the first threshold crossing from all 9,000 trajectories agrees with the logged crossing step in every case. Among Reset failures, X-VLA has 516 cases that never enter this neighborhood and 471 that enter before failing; the corresponding counts are 103 and 384 for $\pi_{0.5}$, and 69 and 243 for HiMem-WAM. This threshold describes approach rather than a verified grasp or necessary task stage: one successful $\pi_{0.5}$ Reset rollout never crosses it.

Among cases where Base and Mid-task succeed but Reset fails, the Reset trajectory never enters the neighborhood in 74/106 X-VLA cases, 21/89 $\pi_{0.5}$ cases, and 2/59 HiMem-WAM cases. This pattern suggests that progress made before the camera change helps the policy approach the object, particularly for X-VLA. Conversely, of the 84 Base/Reset-success cases that fail in Mid-task, all reach a new model query and 13 execute zero old queued actions after the event (eight $\pi_{0.5}$ and five HiMem-WAM cases). Old-action exposure therefore cannot explain every such failure.

\subsection{Milder camera shifts}
\label{sec:camera_magnitude}
To examine whether X-VLA's near-zero Reset success at full strength drives the timing comparison, we evaluate $\alpha\in\{0.25,0.5,1\}$ on the same 1,000 camera cases (700 Plus and 300 PRO). We multiply each case's camera-position, yaw, and field-of-view deltas by $\alpha$, preserving the shift direction and original trigger. These runs retain the checkpoint, initial states, seed, source runtimes, and $H=Q=30$. The design has seven outcomes per case---one shared Base and Reset/Mid-task at each strength---for 7,000 trajectories. Every Mid-task branch retains the exact Base action prefix and recorded pre-event end-effector and trigger-entity positions. The same 916 cases reach the event at every strength, and 915 reach a subsequent model query; all 1,000 remain in the primary denominator.

\begin{figure*}[!t]
    \centering
    \includegraphics[width=\linewidth]{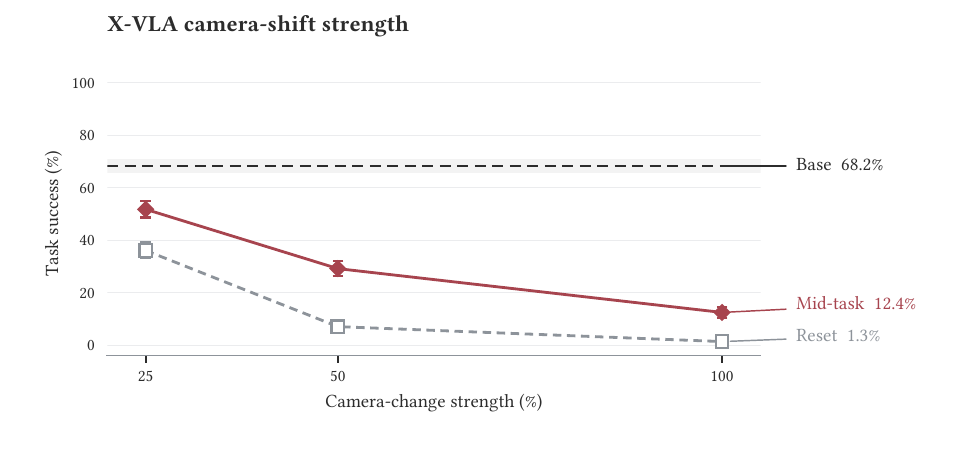}
    \caption{\textbf{Milder camera changes improve X-VLA in both timing conditions.} Reset and Mid-task use 25\%, 50\%, or 100\% of each case's original camera transform. The dashed Base line and pale band show the shared no-event success rate and its interval. Error bars are pointwise 95\% source-stratified empirical bootstrap intervals over the same 1,000 cases. Paired Mid-task minus Reset differences (95\% intervals) are $+15.7$ ($[12.8,18.6]$), $+22.1$ ($[19.3,24.9]$), and $+11.1$ ($[9.2,13.1]$) percentage points at 25\%, 50\%, and 100\% strength, respectively.}
    \label{fig:camera_strength_control}
\end{figure*}

As strength decreases from 100\% to 50\% to 25\%, Reset success rises from 1.3\% to 7.0\% to 36.0\%, and Mid-task success from 12.4\% to 29.1\% to 51.7\% (Fig.~\ref{fig:camera_strength_control}); shared Base success is 68.2\%. At 25\%, 50\%, and 100\% strength, the paired Mid-task minus Reset differences are $+15.7$ points ($[12.8,18.6]$), $+22.1$ ($[19.3,24.9]$), and $+11.1$ ($[9.2,13.1]$), respectively. Mid-task therefore still outperforms Reset at quarter strength, where Reset success is well above its full-strength floor. Both changed-view conditions still fall below Base: Reset by 32.2 points ($[-35.4,-29.0]$) and Mid-task by 16.5 ($[-19.1,-13.9]$). Even this milder camera shift makes the task harder.

The trajectories show whether Reset failures occur before or after approach. At 25\% strength, 119 Reset rollouts fail without entering the 18\,cm neighborhood, 521 enter but fail, and 360 succeed; at 50\%, these counts are 305, 625, and 70. Among cases that succeed in Base and Mid-task but fail in Reset, the Reset trajectory never enters the neighborhood in 18/189 cases at 25\%, 55/221 at 50\%, and 74/106 at full strength. At the milder settings, most of these Reset failures occur after reaching the approach threshold.

Bootstrap resampling keeps all seven outcomes paired within each case and retains the 700/300 source composition. The Mid-task minus Reset gap is largest at 50\% strength: it exceeds the 25\% gap by 6.4 points ($[2.8,10.0]$) and the full-strength gap by 11.0 ($[8.4,13.7]$). The 25\% gap also exceeds the full-strength gap by 4.6 points ($[1.4,7.8]$). These are pointwise paired intervals, rather than simultaneous intervals across strengths and contrasts. The reference-task resampling described above also retains positive Mid-task minus Reset intervals at 25\% and 50\% strength: $[10.9,20.8]$ and $[15.8,28.5]$ points, respectively.

\section{Query Cadence and Action Commitment}
\label{sec:cadence_sweep}

\subsection{Sweep design and complete results}
\label{sec:cadence_design}

The sweep evaluates 18 policy/cadence settings on the fixed 800-pair Lite pool, totaling 14,400 Base/Dynamic pairs. Each setting includes 100 cases per event with a 70:30 Plus/PRO split. Case IDs, event parameters, and policy seeds are fixed across query intervals. Within each setting, Base and Dynamic share an identical executed pre-event prefix. Changing the query interval can change that prefix across settings.

The returned action horizon $H$ and executed prefix $Q$ satisfy $Q\leq H$. The $\pi_{0.5}$ sweep uses $H=10$, distinct from the $H=50$ primary protocol, and tests $Q\in\{2,4,8\}$. The longer intervals are unavailable under that sweep interface. Fast-WAM fixes $H=32$ and tests $Q\in\{2,4,8,12,16\}$. X-VLA and GR00T N1.7 test the same five intervals while decoding $H=Q$ actions. No checkpoint is retrained. For the latter two policies, the decoded horizon and executed prefix vary together, so the sweep compares practical serving settings rather than isolating $Q$ alone.

\begin{table*}[!t]
\caption{\textbf{Complete results for the fixed 800-pair action cadence sweep.} Base and Dynamic are success rates in percent. Gap is Dynamic minus Base in percentage points. $H$ is the decoded action horizon and $Q$ is the number of actions executed before the next observation. Only valid protocols with $Q\leq H$ are listed; $Q\in\{12,16\}$ is therefore undefined for $\pi_{0.5}$ at $H=10$.}
\label{tab:cadence_sweep}
\centering

\setlength{\tabcolsep}{12pt}
\footnotesize
\renewcommand{\arraystretch}{1.10}
\begin{tabular}{@{}l r r r r r@{}}
\toprule
\tableheader Model & \tableheader $Q$ & \tableheader $H$ & \tableheader Base & \tableheader Dynamic & \tableheader Gap \\
\midrule
X-VLA & 2 & 2 & 21.9 & 10.8 & $-11.1$ \\
X-VLA & 4 & 4 & 40.0 & 19.0 & $-21.0$ \\
X-VLA & 8 & 8 & 52.4 & 29.5 & $-22.9$ \\
X-VLA & 12 & 12 & 57.6 & 34.6 & $-23.0$ \\
X-VLA & 16 & 16 & 60.4 & 39.2 & $-21.1$ \\
\midrule
$\pi_{0.5}$ & 2 & 10 & 74.1 & 61.1 & $-13.0$ \\
$\pi_{0.5}$ & 4 & 10 & 77.2 & 64.1 & $-13.1$ \\
$\pi_{0.5}$ & 8 & 10 & 80.0 & 67.2 & $-12.8$ \\
\midrule
GR00T N1.7 & 2 & 2 & 61.2 & 41.9 & $-19.4$ \\
GR00T N1.7 & 4 & 4 & 66.2 & 48.2 & $-18.0$ \\
GR00T N1.7 & 8 & 8 & 69.2 & 51.4 & $-17.9$ \\
GR00T N1.7 & 12 & 12 & 65.5 & 53.2 & $-12.2$ \\
GR00T N1.7 & 16 & 16 & 66.1 & 50.8 & $-15.4$ \\
\midrule
Fast-WAM & 2 & 32 & 38.2 & 18.2 & $-20.0$ \\
Fast-WAM & 4 & 32 & 47.4 & 25.0 & $-22.4$ \\
Fast-WAM & 8 & 32 & 47.6 & 28.0 & $-19.6$ \\
Fast-WAM & 12 & 32 & 46.8 & 26.4 & $-20.4$ \\
Fast-WAM & 16 & 32 & 43.9 & 25.9 & $-18.0$ \\
\bottomrule
\end{tabular}
\end{table*}

X-VLA rises from 21.9\% to 60.4\% Base success and from 10.8\% to 39.2\% Dynamic success as $Q$ increases from 2 to 16. Over its valid range, $\pi_{0.5}$ also improves, with a nearly constant 12.8--13.1-point paired loss. GR00T N1.7 peaks at $Q=8$ in Base and $Q=12$ in Dynamic; Fast-WAM peaks at $Q=8$ in both conditions. Across all settings, Dynamic remains 11.1--23.0 points below Base. These policy-specific optima agree with evidence that fixed chunk optima vary across tasks~\citep{liang2026adaptive}.

\subsection{Native stale-action exposure}
\label{sec:stale_exposure}

For six policies with step-level traces, median stale-action exposure is two actions for $\pi_{0.5}$, four for OpenVLA-OFT and GR00T N1.7, three for VLA-JEPA, and seven for Cosmos-Policy and Fast-WAM. The corresponding means are 2.02, 3.59, 3.60, 2.96, 7.47, and 7.24. The policy can therefore act on stale observations before receiving fresh feedback, even though every response-evaluable case eventually produces a post-event action chunk.

Exposure measures how long feedback is delayed under the native serving protocol. It does not measure internal change detection or physical safety. The sweep shows that query timing changes task success, but changing cadence alone leaves a substantial Dynamic deficit.

\section{Targeted Recovery from Sensor Corruption}
\label{sec:sensor_recovery}

\subsection{Study design and restoration module}
\label{sec:recovery_design}
We test whether a simple observation-processing module can recover part of the
measured performance gap. The study uses X-VLA and 300 fixed
\texttt{sensor\_noise\_onset} cases: 210 from Plus and 90 from PRO, stratified
by source category and corruption setting. Each setting contains 150 cases:
noise standard deviation 24 with 8\% image occlusion, or 36 with 16\%
occlusion. We compare the native policy, always-on restoration, and
quality-gated restoration. Each strategy receives a fresh Base/Dynamic pair
for every case, giving 1,800 rollouts. The released X-VLA checkpoint, policy seed 195, and
native action horizon and query interval $H=Q=30$ are fixed. Within each
strategy, Dynamic replays that strategy's Base actions until the assigned
event; strategies share case IDs and initial physical states, but can have
different executed prefixes. All 300 cases remain in the denominator,
including those that do not reach the trigger. An independent 16-case
development set checks execution and instrumentation. The formal cohort and
restoration parameters were fixed before formal evaluation, without using
its outcomes to select or adjust them. All 900 pairs completed; recorded pre-event actions and
end-effector/trigger-object positions matched within each pair.

\textbf{Observation-only restoration.}
At each policy query, the module independently examines the received
agent-view and wrist RGB images. A high-pass median absolute response on the
green channel estimates noise, excluding inferred occlusions and their
boundaries. The module identifies occlusions as near-black connected components
(all RGB values at most 2) occupying at least 0.2\% of the image, with both
bounding-box sides at least 8 pixels and rectangular fill at least 85\%.
Estimated masks covering more than 30\% of an image are discarded. The gate
activates when estimated noise is at least 12 on the 0--255 intensity scale,
or the inferred mask covers at least 3.5\% of the image. Restoration applies
Telea inpainting with radius 3 pixels, followed by a bilateral filter with
diameter 5, color scale 45, and spatial scale 2 pixels. Always-on restoration
uses the same operator at every query; the gated version passes unflagged
images through unchanged. The module is stateless and requires no training,
clean image, simulator state, event flag, or ground-truth occlusion mask.
It responds to current image quality rather than estimating an event-onset
time.

\subsection{Paired effects}
\label{sec:recovery_effects}
We compare gated restoration with the native policy and use always-on
restoration as an ablation. We report paired 95\% percentile bootstrap intervals from 10,000
resamples, stratified by source category and clustered by task/initial-state
group. The cohort contains 300 distinct groups. For strategy $s$, define gap
closure as $(B_{\mathrm{native}}-D_{\mathrm{native}})-(B_s-D_s)$, where $B$
and $D$ are success rates. We report Base and Dynamic changes separately
so that a drop in Base success cannot be mistaken for recovery.
Table~\ref{tab:sensor_recovery_effects} shows a 7.0-point Dynamic gain and a
6.7-point gap reduction for the gated strategy, with a 0.3-point Base change.
It converts 33 native Dynamic failures to successes while losing 12 native
successes. Always-on restoration also improves Dynamic success. The
0.3-point difference between the two strategies is too uncertain to establish
whether gating improves success over always-on processing. A 19.3-point
Base/Dynamic gap remains under gated restoration.

\begin{table}[!t]
    \centering

    \setlength{\tabcolsep}{7pt}
    \caption{\textbf{Restoration improves Dynamic success with a small aggregate Base change.}
    Paired differences use all 300 X-VLA cases and are in percentage points.
    Gap closure equals Dynamic gain minus Base gain. Intervals are pointwise 95\% bootstrap intervals.}
    \label{tab:sensor_recovery_effects}
    \footnotesize
\renewcommand{\arraystretch}{1.10}
\begin{tabular}{@{}llrc@{}}
        \toprule
\tableheader Comparison & \tableheader Quantity & \tableheader Estimate & \tableheader 95\% CI \\
\midrule
        \multirow[c]{3}{*}{Gated vs. native} & Dynamic gain & $+7.0$ & \textit{[3.0, 11.3]} \\
                        & Base gain    & $+0.3$ & \textit{[$-1.0$, 1.7]} \\
                        & Gap closure  & $+6.7$ & \textit{[2.3, 11.3]} \\
        \midrule
        \multirow[c]{3}{*}{Always vs. native} & Dynamic gain & $+6.7$ & \textit{[2.3, 11.0]} \\
                          & Base gain    & $+1.7$ & \textit{[$-0.7$, 4.0]} \\
                          & Gap closure  & $+5.0$ & \textit{[0.0, 10.0]} \\
        \midrule
        Gated vs. always & Dynamic gain & $+0.3$ & \textit{[$-3.3$, 4.0]} \\
        \bottomrule
    \end{tabular}
\end{table}

\subsection{Coverage and processing cost}
\label{sec:recovery_cost}
Gated restoration increases Dynamic success on both sources and at both
corruption settings (Table~\ref{tab:sensor_recovery_groups}). The intervals
are pointwise; the two corruption settings vary both noise and occlusion.
Gated restoration activates on 1.43\% of measured Base queries and all fresh
Dynamic queries. Replayed prefixes are excluded from these activation
denominators, and Base may already contain source-level visual corruption.
Measured preprocessing of both images takes 9.1\,ms on average in Base and
48.8\,ms in Dynamic (95th percentiles: 9.9 and 58.8\,ms); always-on means are
14.6 and 48.9\,ms. These query-level measurements include quality estimation
and restoration, exclude policy inference and simulation, and describe the
trajectories actually visited. Native diagnostic measurement alone takes
approximately 9--10\,ms, so these are not end-to-end speedup estimates.

\begin{table}[!t]
    \centering

    \setlength{\tabcolsep}{6pt}
    \caption{\textbf{Dynamic gains persist across both sources and corruption settings.}
    Success rates are percentages; gains are percentage points for gated versus native on the same cases.
    Intervals are pointwise 95\% bootstrap intervals. Noise denotes its standard deviation; mask denotes image occlusion.}
    \label{tab:sensor_recovery_groups}
    \footnotesize
\renewcommand{\arraystretch}{1.10}
\begin{tabular}{@{}lrrrrc@{}}
        \toprule
\tableheader Group & \tableheader $N$ & \tableheader Native & \tableheader Gated & \tableheader Gain & \tableheader 95\% CI \\
\midrule
        Plus & 210 & 44.3 & 51.0 & $+6.7$ & \textit{[1.4, 12.4]} \\
        PRO  &  90 & 32.2 & 40.0 & $+7.8$ & \textit{[2.2, 13.3]} \\
        \midrule
        Noise 24, mask 8\%  & 150 & 53.3 & 60.7 & $+7.3$ & \textit{[2.0, 12.7]} \\
        Noise 36, mask 16\% & 150 & 28.0 & 34.7 & $+6.7$ & \textit{[1.3, 12.7]} \\
        \bottomrule
    \end{tabular}
\end{table}

\section{Discussion and Future Directions}
\label{sec:scope_limitations}

\subsection{Interpreting progress on the benchmark}
\label{sec:discussion_interpretation}

The paired gap measures how adding an online event changes task success after the same executed history. A policy must handle both the changed conditions and the need to continue from its current trajectory. Reset controls help distinguish these demands, while paired transitions show whether a higher Dynamic rate preserves existing successes or solves different cases. An effective response should improve post-change completion while retaining Base success. Released policies also differ in training, scale, and action interfaces, so family rankings alone do not identify which design causes greater robustness.

\subsection{Scope and limitations}
\label{sec:discussion_limits}

The benchmark evaluates simulated manipulation using LIBERO's visual, contact, and task abstractions. Its main track introduces one persistent event from a pre-grasp trigger family. This design supports controlled comparisons across event types; transfer to physical robots, later task stages, and repeated or reversible changes requires further evaluation. Geometry screening checks support, workspace, visibility, and collisions, but the human feasibility audit remains incomplete. Some interventions also occur abruptly in the simulator, so their naturalness and physical realization need separate assessment.

The camera controls cover three policies and 1,000 fixed camera cases per policy. The restoration study covers one policy and one sensor-noise event. These studies show how viewpoint difficulty and execution history affect success, and how one restoration module can help. Testing more events, policies, and policy seeds would show how far these findings generalize. Endpoint success and post-event queries measure observable behavior; they do not directly establish internal change detection or deliberate replanning. Safety outcomes were not measured, so task completion cannot establish safe deployment. Measuring contacts and constraint violations would allow future evaluations to assess how safely a policy completes the task.

\subsection{Learning and evaluating responses to change}
\label{sec:future_directions}

Policies could be trained to handle changes during execution. Separately generated dynamic-event trajectories could expose them to altered observations, displaced targets, and recovery behaviors while reserving the frozen benchmark cases for evaluation. Adaptive VLA and WAM policies could learn when to preserve an action sequence, acquire fresh evidence, or revise the plan. Event-aware rewards could favor restored task progress and successful completion while accounting for unnecessary interruptions, additional queries, and unsafe contacts. The benchmark's diagnostics could then guide what these policies are trained to improve.

One proposed mechanism is to compare the next observed representation with a latent state predicted from the preceding observation and action. A task-conditioned gate could use this disagreement to continue execution, cancel the pending action suffix and query again, or stop before replanning. Task relevance matters: a visual change need not invalidate the current action. This remains a future method, to be evaluated against fixed-cadence execution and simpler change triggers under matched observation and computation budgets, including its effects on Base success.

Extensions could also introduce transient, recurrent, and compound event sequences in separate supplemental tracks. Object changes offer a complementary direction to the novel-instance generalization studied by ORION~\citep{zhu_vision_based_manipulation_auro_2026}: a cover could restrict a target's grasp regions, or a detached handle could remove a usable part during execution. Such cases would require stable contact behavior and demonstrated feasibility after the change. Base, reset-change, and mid-task controls could then distinguish changed-object difficulty from execution history, while recovery latency, action cancellation, and contact measurements characterize the response beyond endpoint success.

\end{document}